\documentclass[10pt,twocolumn,letterpaper]{article}

\usepackage[pagenumbers]{cvpr} 

\newcommand{\wvsubmission}{paper}
\newif\ifwvmain
\newif\ifwvappendix
\ifdefstring{\wvsubmission}{paper}{\wvmaintrue}{%
  \ifdefstring{\wvsubmission}{appendix}{\wvappendixtrue}{%
    \ifdefstring{\wvsubmission}{full}{\wvmaintrue\wvappendixtrue}{%
      \PackageError{wvsubmission}{Unknown submission mode `\wvsubmission'}{Use full, paper, or appendix.}%
    }%
  }%
}
\newcommand{\wvprintbibliography}{%
{\small
\bibliographystyle{ieeenat_fullname}
\bibliography{main}}}

\definecolor{cvprblue}{rgb}{0.21,0.49,0.74}
\definecolor{Seattle}{HTML}{00839A}
\usepackage[pagebackref,breaklinks,colorlinks,allcolors=cvprblue]{hyperref}

\def\confName{CVPR}
\def\confYear{2027}

\usepackage{amsmath,amssymb}
\usepackage{booktabs,tabularx,array,colortbl}
\makeatletter
\newcount\wv@col
\newcount\wv@ncols
\newcount\wv@prelen
\newif\ifwv@outerclip
\def\wv@outercliptrue{\global\let\ifwv@outerclip\iftrue}
\def\wv@outerclipfalse{\global\let\ifwv@outerclip\iffalse}
\newif\ifwv@countcols
\def\wv@countcolstrue{\global\let\ifwv@countcols\iftrue}
\def\wv@countcolsfalse{\global\let\ifwv@countcols\iffalse}
\edef\wv@atopen{\detokenize{@{}}}
\def\wv@mark{\wv@mark}
\def\wv@checkpreamble#1{%
  \wv@outerclipfalse
  \edef\wv@pre{\detokenize{#1}}%
  \wv@prelen\z@
  \expandafter\wv@count\wv@pre\wv@mark}
\def\wv@count#1{%
  \ifx#1\wv@mark
    \ifnum\wv@prelen>6
      \expandafter\wv@split\wv@pre\@nil
    \fi
  \else
    \advance\wv@prelen\@ne
    \expandafter\wv@count
  \fi}
\def\wv@split#1#2#3#4\@nil{%
  \def\wv@head{#1#2#3}%
  \ifx\wv@head\wv@atopen
    \wv@lastthree#1#2#3#4\@nil
    \ifx\wv@tail\wv@atopen \wv@outercliptrue \fi
  \fi}
\def\wv@lastthree#1#2#3#4\@nil{%
  \def\wv@rest{#4}%
  \ifx\wv@rest\@empty
    \def\wv@tail{#1#2#3}%
  \else
    \wv@lastthree#2#3#4\@nil
  \fi}
\let\wv@array\@@array
\def\@@array[#1]#2{%
  \wv@checkpreamble{#2}%
  \global\wv@ncols\z@
  \wv@countcolstrue
  \wv@array[#1]{#2}}
\let\wv@mkpream\@mkpream
\def\@mkpream#1{%
  \wv@mkpream{#1}%
  \wv@countcolsfalse}
\let\wv@classz\@classz
\def\@classz{%
  \ifwv@countcols
    \global\advance\wv@ncols\@ne
    \begingroup
      \edef\wv@tmp{%
        \noexpand\@addtopreamble{%
          \noexpand\global\noexpand\wv@col\the\wv@ncols\relax}}%
      \wv@tmp
    \endgroup
  \fi
  \wv@classz}
\def\CT@@do@color{%
  \global\let\CT@do@color\relax
  \@tempdima\wd\z@
  \advance\@tempdima\@tempdimb
  \advance\@tempdima\@tempdimc
  \ifwv@outerclip
    \ifnum\wv@col=\@ne
      \advance\@tempdima-\@tempdimb
      \@tempdimb\z@
    \fi
    \ifnum\wv@col=\wv@ncols
      \advance\@tempdima-\@tempdimc
      \@tempdimc\z@
    \fi
  \fi
  \kern-\@tempdimb
  \leaders\vrule\hskip\@tempdima\@plus 1fill
  \kern-\@tempdimc
  \hskip-\wd\z@ \@plus -1fill }
\makeatother
\usepackage{pifont}
\usepackage{hyperref}
\usepackage{url}
\usepackage{tikz}
\usetikzlibrary{positioning,arrows.meta,calc,shapes.geometric}
\graphicspath{{figures/}}

\newcommand{\icoder}{InteractiveCoder}
\newcommand{\icorpus}{WebVibe-32K}
\newcommand{\webvibe}{WebVibe}
\newcommand{\arena}{WebVibeBench}

\newcommand{\cmark}{\ding{51}}
\newcommand{\xmark}{\ding{55}}

\newcolumntype{Y}{>{\raggedright\arraybackslash}X}

\title{WebVibe: Multimodal Interactive Website Generation via\\
Reverse Synthesis and Visual Self-Distillation}

\author{
Tianxiang Jiang$^{1}$ \quad Sheng Xia$^{4}$ \quad Yilin Long$^{5}$ \quad Kanghui Tian$^{5}$ \quad Linquan Wu$^{6}$ \\
Yongheng Zhang$^{2}$ \quad Daixian Liu$^{7}$ \quad Di Yin$^{2}$ \quad Yi Wang$^{3}$ \quad Yu Qiao$^{3\dagger}$ \quad Qinglong Zhang$^{2\dagger}$ \\
{\small $^{1}$University of Science and Technology of China \quad $^{2}$Tencent Youtu Lab \quad $^{3}$Shanghai AI Lab} \\
{\small $^{4}$Nanjing University \quad $^{5}$Fudan University \quad $^{6}$City University of Hong Kong \quad $^{7}$Tsinghua University} \\
{\small \url{https://github.com/Arsiuuu/WebVibe}}
}

\begin{document}
\ifwvmain
\twocolumn[{%
\renewcommand\twocolumn[1][]{#1}%
\maketitle
\begin{center}
\captionsetup{type=figure}
\vspace{-6mm}
\includegraphics[width=\textwidth]{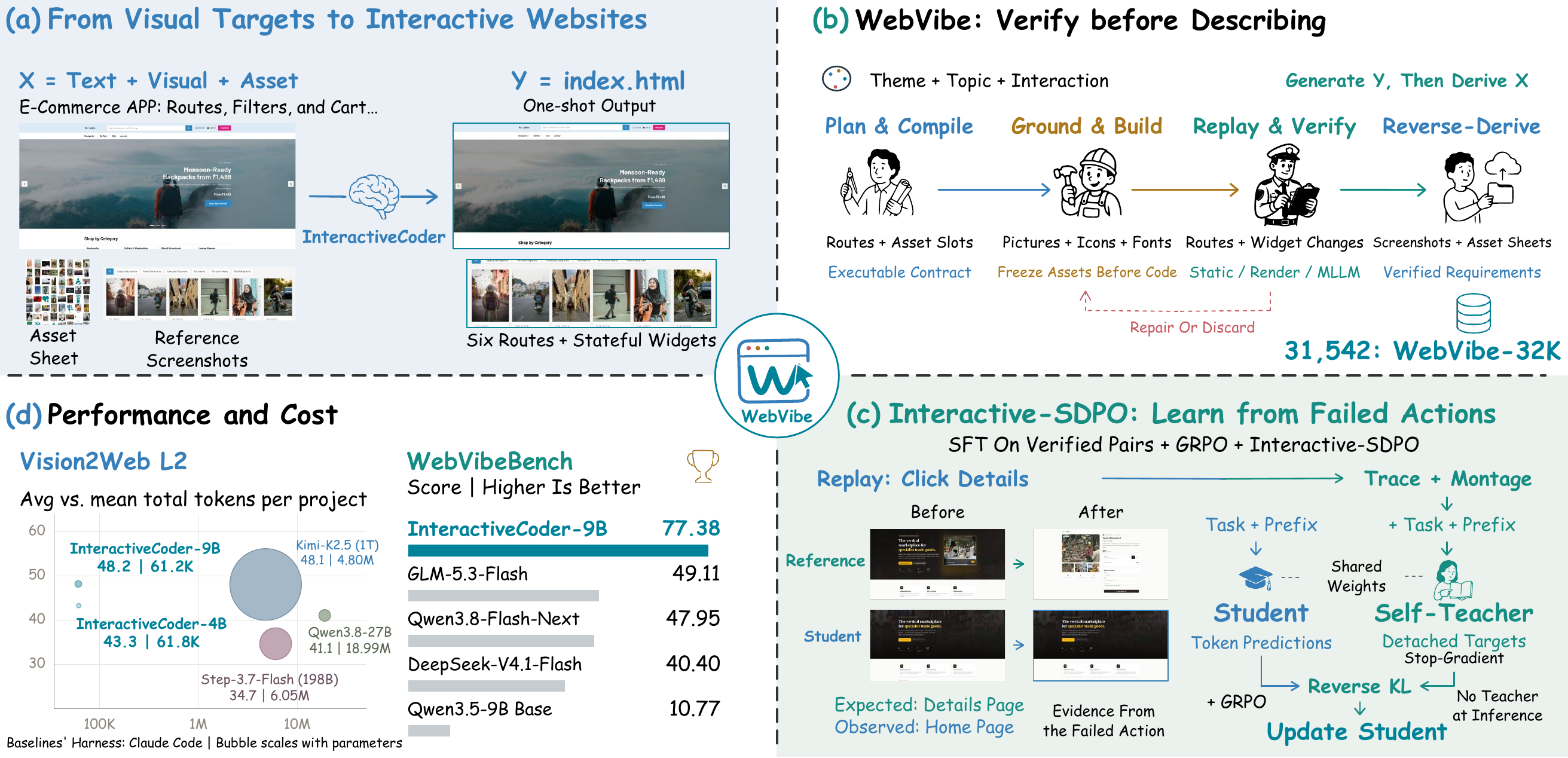}
\vspace{-6mm}
\captionof{figure}{\textbf{Overview of \webvibe{}.}
(a)~The Vision2Web task requires a model to generate a multi-page interactive website $Y$ from a task description, route screenshots, and visual asset sheets $X=(\text{Text},\,\text{Visual},\,\text{Asset})$.
(b)~Our data synthesis pipeline \webvibe{} reverses synthesis ($Y\to X$): it verifies every compiled route and widget check in a headless browser before deriving $X$ from replay-confirmed states, yielding the \icorpus{} corpus (Section~\ref{sec:dp}).
(c)~During training, Interactive-SDPO pairs GRPO~\cite{shao2024deepseekmath} with a weight-sharing self-teacher conditioned on the interaction trajectory and a $2\times2$ state-transition montage, injecting interaction-grounded visual feedback as privileged context to guide learning (Section~\ref{sec:learning}).
(d)~Our \icoder{}-4B and \icoder{}-9B models outperform all evaluated baselines on \arena{} and match multi-turn coding agents on Vision2Web L2~\cite{he2026vision2web} with far fewer tokens and parameters (Section~\ref{sec:experiments}).}
\label{fig:teaser}
\end{center}
}]

{
  \renewcommand{\thefootnote}{}
  \footnotetext{$^{\dagger}$Corresponding authors.}
}


\begin{abstract}
Although multimodal large language models translate single screenshots into static HTML, generating interactive functional websites from design specifications still struggles with multi-page routing and stateful widget transitions. Examining existing training pipelines reveals two bottlenecks: (1)~in forward data synthesis ($X\to Y$), writing the specification $X$ before code execution leaves unimplemented or altered requirements in $X$ as supervision noise; and (2)~in reinforcement learning, scalar rewards discard the rich state transitions observed during replay. Based on these findings, we introduce \textbf{\webvibe{}} and \textbf{Interactive-SDPO} to train \textbf{\icoder{}}. \webvibe{} reverses synthesis into $Y\to X$ by verifying cross-page and in-page interactions in the browser before deriving multimodal inputs from replay-confirmed states to build the 31{,}542-site \textbf{\icorpus{}} corpus with reusable execution contracts. Interactive-SDPO then conditions a weight-sharing self-teacher on the replay trace and a $2\times2$ state-transition montage to provide token-level supervision at zero inference-time overhead. On the public Vision2Web L2, one-shot \icoder{}-9B nearly \textbf{3$\boldsymbol{\times}$} its base model's functional score (20.2 to \textbf{55.2} FS) and matches a 1T-parameter coding agent in a single pass with 111$\times$ fewer total parameters and 78$\times$ fewer tokens. On our \arena{}, \icoder{}-4B and 9B further outperform all evaluated baselines by over \textbf{23} points.
\end{abstract}

\vspace{-4mm}
\section{Introduction}
\label{sec:introduction}

Reconstructing an interactive functional website from visual designs requires more than reproducing static pixels. A practical frontend specification pairs route screenshots and local visual assets with a brief behavior description~\cite{he2026vision2web,chen2026iwr}. Turning it into a working application requires not only faithfully recovering layouts and grounding local assets, but also accurately implementing cross-page and in-page interactions such as route navigation and stateful widget updates. Mastering these interactions is the core challenge of interactive website generation, as they never appear in one image.

Previous large-scale corpora such as WebSight~\cite{laurenccon2024unlocking} and WebCode2M~\cite{gui2025webcode2m} enable open models to reproduce static single-page appearance. However, visually reproducing multi-page interactive websites remains difficult due to bottlenecks in both training data construction (Table~\ref{tab:corpora}) and reinforcement learning supervision. On the data side, crawling live websites~\cite{gui2025webcode2m,si2025design2code} rarely yields self-contained code, local asset bundles, or executable test contracts, while existing synthetic pipelines either rely on text-only instructions~\cite{lu2025webgenbenchevaluatingllmsgenerating} or remain single-page without local visual assets~\cite{xu2025webvia,xu2026video2code,xiao2026interwebcoder}. More fundamentally, forward synthesis ($X\to Y$) fixes the specification $X$ before executing code $Y$; when generated routes, widget states, or repaired DOM labels diverge from the initial plan, unimplemented requirements remain in $X$ as supervision noise.

Interactive behavior can be verified only by executing the program. Our \textbf{\webvibe{}} engine therefore reverses synthesis into $Y\to X$ around headless browser replay (Figure~\ref{fig:teaser}b). Before code generation, \webvibe{} compiles a planned site into hash routes, MLLM-verified asset slots, and an executable contract $\mathcal{C}_X$. Non-intrusive browser replay accepts a generated site $Y$ only when it confirms through human-visible controls that (1)~every planned navigation reaches its target route and retraces under the Back button, and (2)~every in-page widget satisfies semantic state-flip assertions rather than merely clicking without error as in RILA~\cite{guo2026rendering}. Only after $Y$ passes verification does the engine derive $X=(\text{Text},\,\text{Visual},\,\text{Asset})$ from replay-confirmed states to build \textbf{\icorpus{}}, a corpus of 31{,}542 verified multi-page websites with reusable verification environments $(\mathcal{C}_X,\mathcal{P}_X)$.

\begin{table*}[t]
\caption{\textbf{Comparison of UI-to-code training  datasets.}
WebVibe-32K pairs multimodal inputs (Text specification, Visual page screenshots and local Asset sheets) with multi-page routing and browser-replayed state transitions.}
\label{tab:corpora}
\centering
\footnotesize
\setlength{\tabcolsep}{4.5pt}
\begin{tabularx}{0.9\textwidth}{@{}>{\raggedright\arraybackslash}X c c c r c c c@{}}
\toprule
 & \multicolumn{3}{c}{Input} & & & & \\
\cmidrule(lr){2-4}
Dataset & Text & Visual & Asset & Samples & Multi-page & Interaction & Verification \\
\midrule
\multicolumn{8}{@{}l}{\textit{Static screenshot-to-code}} \\
Pix2Code~\cite{beltramelli2018pix2code}                          & \xmark & \cmark & \xmark & 1.7K  & \xmark & \xmark     & \xmark \\
WebSight v0.2~\cite{laurenccon2024unlocking}                     & \xmark & \cmark & \xmark & 1.92M & \xmark & \xmark     & Render + Content filter \\
WebCode2M~\cite{gui2025webcode2m}                                & \xmark & \cmark & \xmark & 2.56M & \xmark & \xmark     & Render + Scorer \\
Web2Code~\cite{yun2024web2code}                                  & \cmark & \cmark & \xmark & 885K  & \xmark & \xmark     & Render + LLM refinement \\
Screen-10K (ScreenCoder)~\cite{jiang2026screencoder}             & \xmark & \cmark & \xmark & 10K   & \xmark & \xmark     & Automatic filtering \\
\midrule
\multicolumn{8}{@{}l}{\textit{Multi-page or interactive}} \\
WebGen-Instruct~\cite{lu2025webgenbenchevaluatingllmsgenerating} & \cmark & \xmark & \xmark & 6.7K & \cmark & \cmark & Appearance score \\
WebView (WebVIA)~\cite{xu2025webvia}                            & \cmark & \cmark & \xmark & 11K   & \xmark & \cmark     & Render + Human spot-check \\
WebVidCoding (Video2Code)~\cite{xu2026video2code}                & \xmark & \cmark & \xmark & 12K   & \xmark & \cmark     & Render + Consistency checks \\
InterWebCoder~\cite{xiao2026interwebcoder}                      & \xmark & \cmark & \xmark & 1.7K  & \xmark & \cmark     & Tests + State similarity \\
\midrule
\rowcolor{cvprblue!8}
\textbf{\icorpus{} (ours)}                                     & \cmark & \cmark & \cmark & \textbf{31.5K} & \cmark & \cmark & \textbf{Replayed state change} \\
\bottomrule
\end{tabularx}
\vspace{-4mm}
\end{table*}

On the training side, supervised fine-tuning on \icorpus{} teaches the policy to write multi-page React applications, yet token-level imitation can fall into repetitive loops. Rather than relying on slow and unstable GUI-agent rollouts for execution feedback~\cite{lu2025webgenagentenhancinginteractivewebsite}, Stage~2 reinforcement learning reuses each site's compiled contract $\mathcal{C}_X$ and reference states $\mathcal{P}_X$ as a fast, deterministic browser verifier. Optimizing a gated execution reward over functional replay $F(Y)$ and visual similarity $V(Y)$ with GRPO~\cite{shao2024deepseekmath} penalizes truncated or unrenderable code to boost delivery. Still, browser replay records which click reached the wrong route and which filter left a list unchanged, whereas GRPO compresses the entire trajectory into one scalar. \textbf{Interactive-SDPO} returns this transition evidence to training. For the lowest-reward failed rollouts, a weight-sharing self-teacher receives the replay trace and a $2\times2$ state-transition montage around the first failed action as privileged visual context, distilling token-level supervision into the student via reverse KL alongside GRPO at zero inference overhead.

This two-stage recipe yields \textbf{\icoder{}-4B} and \textbf{\icoder{}-9B}, turning compact backbones into strong one-shot website generators across interactive frontend benchmarks (Figure~\ref{fig:teaser}d). On Vision2Web L2~\cite{he2026vision2web}, \icoder{}-9B nearly triples its base functional score (20.2 to \textbf{55.2}, the highest one-shot FS; \textbf{48.2} Avg), outperforming GPT-5.4~\cite{openai2026gpt54} and Qwen3.8-Flash-Next~\cite{qwen3.8flashnext} while matching the 1T-parameter Kimi-K2.5~\cite{team2026kimi} coding agent with 111$\times$ fewer parameters and 78$\times$ fewer tokens; \icoder{}-4B likewise outperforms the 7$\times$ larger Qwen3.8-27B~\cite{qwen38} in both one-shot and agentic settings. 
On \arena{}, our 200-site in-distribution benchmark across 87 domains, \icoder{}-4B and 9B reach \textbf{72.89} and \textbf{77.38}, surpassing frontier baselines such as GLM-5.3-Flash~\cite{zeng2026glm} and DeepSeek-V4.1-Flash~\cite{deepseekai2026deepseekv41flash} by more than \textbf{23} points. Under a matched RL budget, Interactive-SDPO improves over GRPO by \textbf{+7.12} Function Score on Vision2Web.

In summary, our contributions are threefold:
\begin{itemize}[leftmargin=*,itemsep=1pt,topsep=2pt]
\item \textbf{\icorpus{}.} A 31{,}542-site corpus with visual asset contact sheets, reusable execution contracts, and the 200-site \arena{}, built by our \webvibe{} reverse synthesis ($Y\to X$) and browser-replay verification engine.
\item \textbf{Interactive-SDPO.} Transition-based self-distillation that pairs browser replay interaction traces with $2\times2$ state-transition montages as privileged visual context for token-level RL supervision at zero inference-time overhead.
\item \textbf{\icoder{}.} Compact 4B and 9B one-shot generators achieving strong accuracy on Vision2Web and \arena{} at a fraction of the parameter and token cost.
\end{itemize}

\section{Related Work}
\label{sec:related}

\paragraph{Multimodal LLMs for Code Generation.}
Large language models (LLMs) have made substantial progress in text-to-code synthesis, yet many real programming tasks state their intent through visual artifacts such as screenshots~\cite{yun2024web2code}, slides~\cite{ge2025autopresent}, charts~\cite{tang2026mm}, vector graphics~\cite{xing2025empowering}, code-driven video~\cite{diao2026vibevideobenchevaluatingmultimodalagents}, 3D worlds~\cite{lu2026worldcoder,fang2026blenderworld}, and 4D scene previs~\cite{chen2026vibeprevisbenchevaluatingagenticcoding}.
Multimodal LLMs~\cite{jiang2025vknowu,jiang2026imagine} now take these artifacts as input and write the program behind them~\cite{sun2026januscoder,zhao2025vincicoder,zhao2026beyond}. For webpages, WebCode2M~\cite{gui2025webcode2m} pairs static rendered pages with HTML at scale, while Design2Code~\cite{si2025design2code}, ScreenCoder~\cite{jiang2026screencoder}, Widget2Code~\cite{zhang2026widget2code}, and MRWeb~\cite{wan2024mrweb} advance layout, widget, and multi-page resource-aware reconstruction. Moving beyond static appearance to interactive websites, Interaction2Code~\cite{xiao2025interaction2code} and Vision2Web~\cite{he2026vision2web} specify target behavior with UI prototypes and requirement documents, while IWR-Bench~\cite{chen2026iwr} and WebVR~\cite{dai2026webvr} use interaction videos. On the generation side, WebVIA~\cite{xu2025webvia} collects multi-state screenshots with an exploring agent, and Video2Code~\cite{xu2026video2code} revisits action-critical video segments at higher temporal resolution. Meanwhile, UI2Code$^N$~\cite{yang2025ui2code}, RILA~\cite{guo2026rendering}, and WebCoder~\cite{dou2026webcoder} rely on iterative visual polishing, test-time replay revision, or multi-turn agentic repair over evolving project states. 
WebSight~\cite{laurenccon2024unlocking} already renders screenshots from generated code, and backtranslation~\cite{li2024selfalignmentinstructionbacktranslation,wu2024inversecoderselfimprovinginstructiontunedcode} derives instructions from existing outputs, but neither verifies multi-page interactive execution. \webvibe{} instead derives multimodal inputs ($\text{Text}$, route screenshots, and anonymized asset sheets) strictly from browser-confirmed interactive states alongside reusable execution contracts, enabling \icoder{} to generate multi-page interactive websites in a single pass.

\begin{figure*}[t]
\centering
\includegraphics[width=\textwidth]{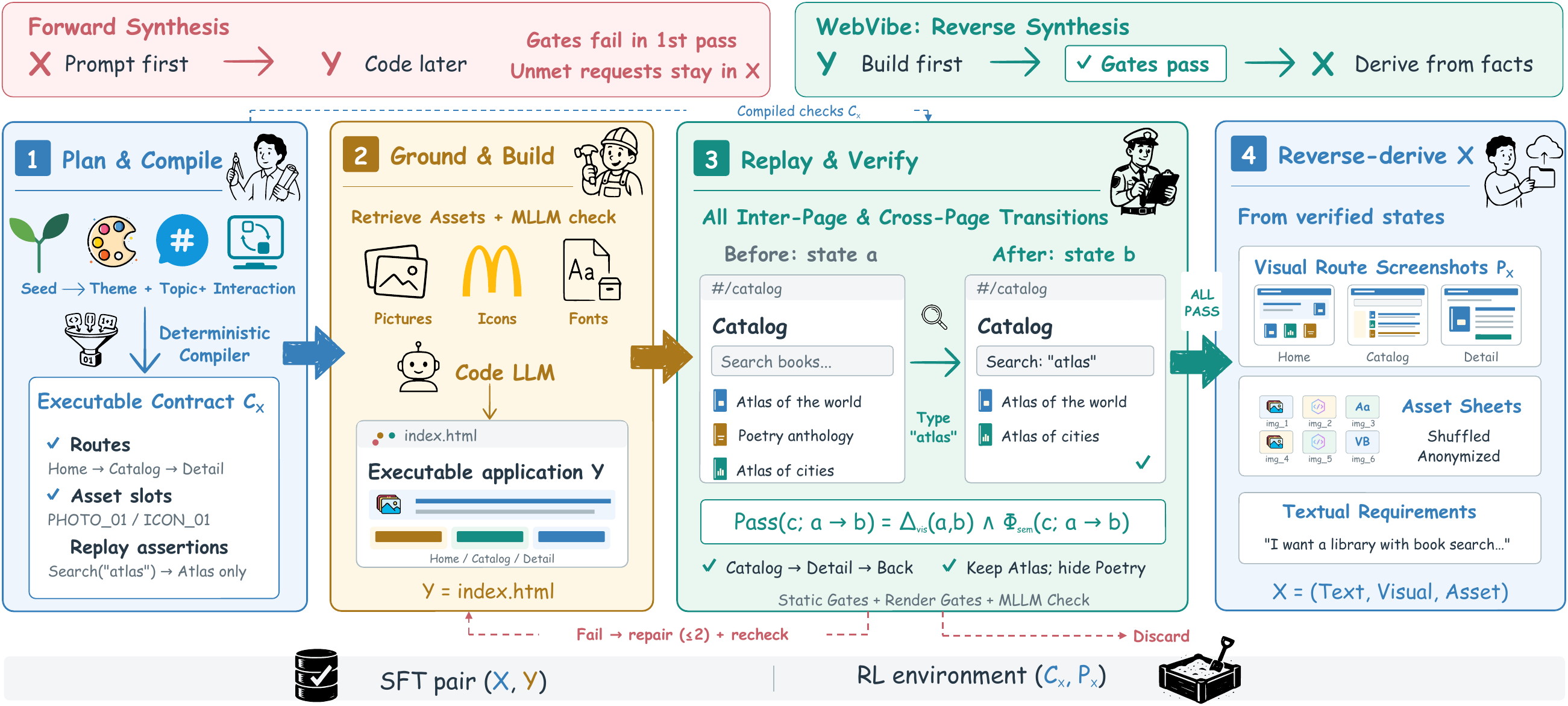}
\caption{\textbf{The \webvibe{} reverse synthesis engine ($Y\to X$).}
A seeded controller and an LLM planner compile an executable contract, then an MLLM grounds the visual assets, and a code LLM synthesizes the site. Static, browser-replay, and render gates verify or repair the page before the task description is derived from replay-confirmed states.}
\label{fig:dp-overview}
\vspace{-4mm}
\end{figure*}

\vspace{-3mm}
\paragraph{Reinforcement Learning and Self-Distillation.}
Execution-guided reinforcement learning optimizes code generators from runtime feedback. Early methods~\cite{le2022coderl,shojaee2023execution,gehring2024rlef} rely on PPO or a value critic over compiler and test feedback, while GRPO~\cite{shao2024deepseekmath,zhu2024deepseek} removes the critic by normalizing rewards within sampled rollout groups. Building on it, SWE-RL~\cite{wei2025swerladvancingllmreasoning} trains on software evolution data, and WebGen-Agent~\cite{lu2025webgenagentenhancinginteractivewebsite} and InterWebCoder~\cite{xiao2026interwebcoder} optimize GUI-agent or executable-test rewards for interactive website generation. However, compressing browser execution into a single scalar reward discards the state transitions behind each failure. On-policy distillation~\cite{agarwal2024policy} recovers token-level supervision via reverse KL to a teacher, and recent self-distillation methods~\cite{zhao2026self,yang2026self} reuse the student as a self-teacher conditioned on privileged context.
SDPO~\cite{hubotter2026reinforcement} feeds runtime errors or passing rollouts, and visual self-distillation methods~\cite{li2026visual,bi2026opd,liang2026visual} extend this paradigm to cross-modal reasoning and static visual observations such as rendered charts and webpages~\cite{dong2026self}. Interactive-SDPO instead injects interactive visual feedback as a privileged context, conditioning the weight-sharing self-teacher on the replay interaction trace and a state-transition montage around the first failed action.

\section{WebVibe Construction}
\label{sec:dp}
\label{sec:dataset}
\label{sec:pipeline}

To prevent unimplemented behaviors from entering the training prompt $X$, \webvibe{} reverses data synthesis into $Y \to X$ (Figure~\ref{fig:dp-overview}): it compiles a site plan into an executable contract (Section~\ref{sec:dp-format}), grounds visual assets and verifies every route and widget transition in a headless browser (Section~\ref{sec:dp-verify}), and derives the student input $X$ strictly from replay-confirmed states (Section~\ref{sec:dp-pack}). Every accepted site yields both a verified supervised pair $(X,Y)$ and a reusable verification environment $(\mathcal{C}_X,\mathcal{P}_X)$ for reinforcement learning. 

\subsection{Task Formulation and Contract Compilation}
\label{sec:dp-format}
\label{sec:dp-compile}

We formulate interactive website reconstruction as a single-turn multimodal program synthesis task,
\begin{equation}
  X=\bigl(\text{Text},\;\text{Visual},\;\text{Asset}\bigr),\qquad Y=\text{\texttt{index.html}},
  \label{eq:dp-sample}
\end{equation}
where $\text{Text}$ is a natural-language task description, $\text{Visual}$ is the ordered set of full-page route screenshots, and $\text{Asset}$ is a set of visual contact sheets rendering all candidate local assets. The target $Y$ is a self-contained single-file React application with zero external network dependencies. Because the contact sheets pack candidate images, icons, and font specimens alongside their bare filenames into compact grids, the model grounds local assets visually in a single pass without slow multi-turn agentic file inspection.

To prevent the template homogeneity typical of LLM-generated webpages, synthesis begins from a \emph{style card} distilled offline from a real website, specifying visual attributes (palette, typography, and layout density), paired with a sampled domain and structural budget. Conditioned on this seed, an LLM planner proposes a structured site specification covering pages, data collections with detail views, navigation links, and interactive widgets (e.g., tabs, filters, search, carousels, and forms) with explicit control labels and expected outcomes. Rather than trusting the LLM to write both code and tests, a deterministic compiler validates structural reachability and compiles the plan into hash routes, typed asset slots, and an executable contract $\mathcal{C}_X$ spanning every planned navigation journey and widget transition. The code generator receives only the resolved routes and asset slots, withholding $\mathcal{C}_X$ for browser verification.

\subsection{Synthesis and Replay Verification}
\label{sec:dp-verify}

Synthesizing code before selecting visual assets often leads to broken paths or semantic mismatches between text and imagery. \webvibe{} therefore freezes all visual assets before code generation: candidate photos are retrieved by semantic tags from a 10M-image Pexels index, checked for subject alignment by an MLLM validator that replaces off-topic picks, and bundled with matching icons, SVG wordmarks, and assigned webfonts. Conditioned on the compiled plan and frozen assets, code LLMs synthesize the application~$Y$.

Because a rendered page can still route to dead links or mishandle user actions, a headless browser runs every check $c\in\mathcal{C}_X$ via human-visible control labels and page regions (\texttt{header}, \texttt{main}, \texttt{footer}) after static checks on runtime errors and asset integrity. A navigation step passes only if clicking the visible control switches the route, renders the expected destination view and heading, and retraces cleanly under the browser Back button. Across the six in-page widget families (filters, search, tabs, accordions, carousels, and forms), a superficial DOM or pixel change does not guarantee that the intended logic ran. Verification thus enforces \emph{non-trivial state change} and \emph{semantic assertions}:
\begin{equation}
  \operatorname{Pass}(c;\,a\to b) = \Delta_{\mathrm{vis}}(a,b)\;\wedge\;\Phi_{\mathrm{sem}}(c;\,a\to b),
  \label{eq:dp-change}
\end{equation}
where $\Delta_{\mathrm{vis}}(a,b)\in\{0,1\}$ tests whether the main-content DOM text or the masked viewport around a control changes from state $a$ to state $b$, ruling out trivial self-echoing updates such as a button toggling its own highlight. Beyond change detection, $\Phi_{\mathrm{sem}}(c;\,a\to b)\in\{0,1\}$ enforces the compiled state-flip assertions in $\mathcal{C}_X$, requiring target items, panels, or submission states to appear and excluded items to hide.


\begin{figure*}[t]
\centering
\begin{minipage}[t]{0.49\textwidth}\centering
\includegraphics[width=\linewidth]{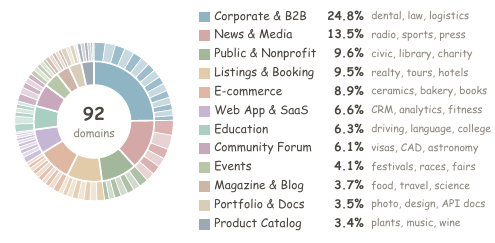}\\[1pt]
{\small (a) Site archetypes and their 92 domains}
\end{minipage}\hfill
\begin{minipage}[t]{0.49\textwidth}\centering
\includegraphics[width=\linewidth]{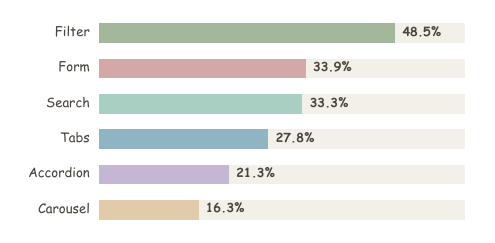}\\[1pt]
{\small (b) Sites with each verified in-page widget}
\end{minipage}
\caption{\textbf{Domain and widget distributions of \icorpus{}} across 12 archetype groups, 92 domains, and six widget types.}
\label{fig:corpus-stats}
\vspace{-2mm}
\end{figure*}

\subsection{Reverse Input Derivation}
\label{sec:dp-pack}
\label{sec:dp-stats}

Because code synthesis and repair alter actual DOM copy, control labels, and layouts relative to the initial plan, all three components of $X$ are derived from the final verified site~$Y$ so that $\text{Text}$, $\text{Visual}$, and $\text{Asset}$ remain strictly aligned. Full-page screenshots ($\text{Visual}$) are captured across the planned routes of~$Y$, while route and post-action captures are saved as reference states $\mathcal{P}_X$. A text-only writer then generates the owner-perspective description ($\text{Text}$) strictly from DOM facts and only the links, journeys, and widgets confirmed by browser replay; any mention of implementation internals or unverified control labels is rejected. Finally, the packer adds distractor images, icons, and wordmarks to the asset folder, shuffles and anonymizes all files (e.g., \texttt{img\_7.jpg}), and renders the contact sheets ($\text{Asset}$) so the student policy grounds assets visually.

Running the pipeline yields \icorpus{}, a corpus of 31{,}542 verified multi-page interactive websites across 92 domains (Figure~\ref{fig:corpus-stats}) paired with executable contracts $\mathcal{C}_X$ and reference states $\mathcal{P}_X$ for reinforcement learning. Using isolated seeds, site plans, and deduplicated assets, the same pipeline constructs \textbf{\arena{}}, a held-out benchmark of 200 multi-page interactive websites across 87 domains.

\section{Method}
\label{sec:learning}
\label{sec:method}

We train the base MLLM policy $\pi_\theta$ in two stages: Stage~1 performs supervised fine-tuning (SFT) on \icorpus{} to learn multi-page layout, routing syntax, and visual asset grounding, while Stage~2 applies browser-in-the-loop reinforcement learning with Interactive-SDPO (Figure~\ref{fig:teaser}c).

\subsection{Supervised Fine-Tuning}
\label{sec:prelim}
\label{sec:sft}

Given a multimodal specification $X$ from Eq.~\eqref{eq:dp-sample}, the policy $\pi_\theta$ generates a candidate application $Y=(y_1,\ldots,y_T)$. Stage~1 initializes $\pi_\theta$ on \icorpus{} by minimizing the standard autoregressive negative log-likelihood over the $T$ tokens of each verified application~$Y$:
\begin{equation}
  \mathcal{L}_{\text{SFT}}(\theta)=-\,\mathbb{E}_{(X,Y)}
  \sum_{j=1}^{T}\log\pi_\theta\bigl(y_j\mid X,\,y_{<j}\bigr).
  \label{eq:sft}
\end{equation}
While SFT teaches the policy to emit complete multi-page React applications and ground visual assets, token-level imitation does not verify whether generated routes and widgets execute without runtime or behavioral errors, motivating Stage~2 reinforcement learning with browser replay.

\subsection{GRPO with Gated Execution Reward}
\label{sec:setting}
\label{sec:reward}
\label{sec:rl}

During reinforcement learning, executing the compiled contract $\mathcal{C}_X$ on a rollout $Y$ in a headless browser yields three signals: (1)~a binary render gate $g(Y)\in\{0,1\}$ checking that $Y$ is complete (non-truncated), mounts in React without runtime errors, and renders visible content; (2)~a functional score $F(Y)\in[0,1]$, the fraction of navigation and widget checks $c\in\mathcal{C}_X$ passing $\operatorname{Pass}(c;\,a\to b)$ in Eq.~\eqref{eq:dp-change}; and (3)~a visual score $V(Y)\in[0,1]$, the mean block-wise visual similarity from a fixed MLLM judge across reference states in $\mathcal{P}_X$, scoring $0$ on unreached states. These signals form the gated execution reward:
\begin{equation}
  R(Y)=g(Y)\,\bigl(0.5\,F(Y)+0.5\,V(Y)\bigr),
  \label{eq:reward}
\end{equation}
which assigns zero reward to unrenderable code while balancing functional correctness and visual fidelity.

Given a prompt $X$, $\pi_{\mathrm{old}}$ samples $G$ rollouts $\{Y_i\}_{i=1}^G$ with rewards $\{R_i\}_{i=1}^G$ and group-standardized advantages $\{\hat A_i\}_{i=1}^G$. GRPO~\cite{shao2024deepseekmath} maximizes the clipped objective:
\begin{equation}
\begin{aligned}
  \mathcal{J}_{\mathrm{GRPO}}(\theta) &= \mathbb{E}\Bigl[\tfrac{1}{N}\textstyle\sum_{i=1}^{G}\sum_{j=1}^{T_i}
  \min\bigl(\rho_{i,j}\hat A_i,\\[-2pt]
  &\qquad\quad\;\operatorname{clip}(\rho_{i,j},1-\epsilon,1+\epsilon)\,\hat A_i\bigr)\Bigr],
\end{aligned}
  \label{eq:objective}
\end{equation}
where $N=\sum_{i=1}^{G}T_i$ is the group token count and $\rho_{i,j}$ is the importance ratio between new and old token probabilities.

\subsection{Interactive-SDPO}
\label{sec:interactive-sdpo}

Compressing a multi-step browser replay into one scalar $R_i$ discards the state transitions behind a failure and broadcasts $\hat A_i$ uniformly across tens of thousands of code tokens. Moreover, because interactive bugs lie in action-triggered state transitions ($a\to b$) rather than a single static view, neither text-only code self-distillation~\cite{hubotter2026reinforcement} nor static single-image visual self-distillation~\cite{dong2026self} can diagnose whether a click switched routes, a filter updated its list, or a form completed submission. \mbox{Interactive-SDPO} therefore injects \emph{interactive visual feedback} from browser replay as privileged context into a weight-sharing self-teacher, turning dynamic state discrepancies into dense token-level supervision on the student's own rollouts.

\begin{figure}[t]
\centering
\includegraphics[width=\linewidth]{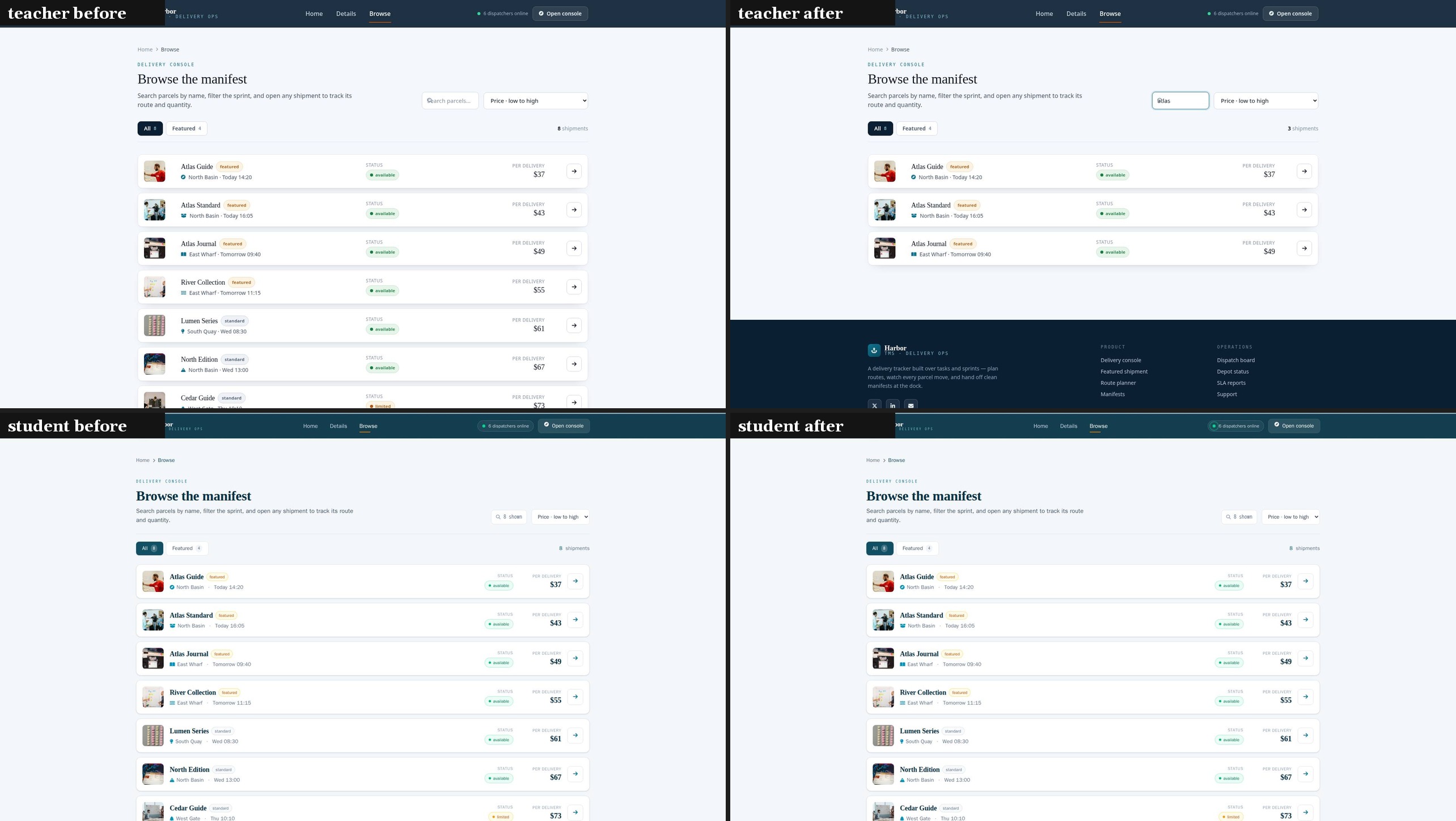}
\caption{\textbf{State-transition montage $M_i$ at a failed search check.}
The query filters the reference (“teacher” in the prompt) from 8 to 3 rows (top); replay cannot find the student's search field by its label, so 8 rows remain (bottom).}
\label{fig:sdpo-montage}
\vspace{-4mm}
\end{figure}

\paragraph{Interactive Visual Feedback.}
From the replay of rollout $Y_i$, we extract privileged feedback $h_i=(\tau_i,M_i)$. The trace $\tau_i$ logs ordered action intents, pass/fail outcomes, runtime errors, and expected versus observed DOM text deltas up to the first failed action $k_i$. We pair it with a $2\times2$ state-transition montage $M_i$ (Figure~\ref{fig:sdpo-montage}) stacking the reference and candidate viewports immediately before and after action~$k_i$. When $Y_i$ fails the render gate ($g(Y_i)=0$), $h_i$ supplies only the sanitized runtime error log.
\paragraph{Weight-Sharing Self-Teacher.}
Injecting $h_i$ before the assistant response turns the policy $\pi_\theta$ itself into a self-teacher that re-scores the student's trajectory without a separate teacher network or reference code:
\begin{equation}
\begin{aligned}
  s_{i,j}&=\pi_\theta(\cdot\mid X,y_{i,<j}),\\
  t_{i,j}&=\operatorname{sg}
    \bigl[\pi_\theta(\cdot\mid X,h_i,y_{i,<j})\bigr],
\end{aligned}
  \label{eq:isdpo-teacher}
\end{equation}
where $\operatorname{sg}[\cdot]$ denotes the stop-gradient operator. Over a subset $\mathcal{S}_X\subseteq\{1,\ldots,G\}$ of defective rollouts, we minimize the full-vocabulary reverse KL divergence:
\begin{equation}
  \mathcal{L}_{\mathrm{I\text{-}SDPO}}(\theta)\!=\!
  \mathbb{E}\!\left[\frac{1}{N_{\mathcal{S}}}\!
  \sum_{i\in\mathcal{S}_X}\!\sum_{j=1}^{T_i}
  w_{i,j}D_{\mathrm{KL}}(s_{i,j}\Vert t_{i,j})\right]\!,
  \label{eq:isdpo-kl}
\end{equation}
where $N_{\mathcal{S}}=\sum_{i\in\mathcal{S}_X}T_i$ is the distilled token count and $w_{i,j}=\operatorname{sg}[\min(2,\rho_{i,j})]$ bounds policy drift from $\pi_{\mathrm{old}}$.

\subsection{Training Objective and Inference}
\label{sec:objective}
\label{sec:modes}

All $G$ rollouts participate in the GRPO update, whereas distillation focuses compute on $\mathcal{S}_X$: from the defective rollouts $\mathcal{E}_X\subseteq\{1,\ldots,G\}$ with valid replay diagnostics, we select the $\min(K,|\mathcal{E}_X|)$ lowest-reward candidates. We then jointly optimize the sequence-level GRPO objective and the token-level self-distillation loss ($\lambda$ balances transition-guided distillation and GRPO reward maximization):
\begin{equation}
  \mathcal{L}(\theta)=
  -\mathcal{J}_{\mathrm{GRPO}}(\theta)
  +\lambda\,\mathcal{L}_{\mathrm{I\text{-}SDPO}}(\theta),
  \label{eq:isdpo-objective}
\end{equation}
At inference time, browser replay, the transition montage, and the teacher branch are discarded: the policy takes only the task input $X$ and synthesizes $Y$ in a single forward pass.

\section{Experiments}
\label{sec:experiments}


We evaluate on two interactive frontend benchmarks. \textbf{Vision2Web L2}~\cite{he2026vision2web} comprises 66 multi-page projects specified by UI prototypes and requirement documents, reporting Functional Score (FS) from GUI-agent workflows, block-wise Visual Score (VS), their average (Avg), and Delivery Success Rate (DSR), with undelivered projects scored as $0$. \textbf{\arena{}} covers 200 multi-page interactive websites across 87 domains, disjoint from the SFT and RL datasets; it reports the render gate rate (Gate), functional replay ($F$), visual similarity ($V$), and overall Score (Eq.~\eqref{eq:reward}); 

\icoder{}-4B/9B are initialized from Qwen3.5-4B and Qwen3.5-9B~\cite{qwen3.5}. Stage~1 fine-tunes them on \icorpus{} for two epochs, and Stage~2 trains with Interactive-SDPO on contract-verified prompts using Qwen3.8-27B~\cite{qwen38} as the visual reward judge. We compare against one-shot frontier and open-weight MLLMs under a 65{,}536-token budget, as well as multi-turn coding agents with the Claude Code harness.

\begin{table}[t]
\caption{\textbf{Main results on Vision2Web L2}.
\emph{Agent (+ Claude Code)}: multi-turn agentic generation, from the Vision2Web~\cite{he2026vision2web} paper. 
\emph{One-shot}: single-pass. 
Our models are in \textbf{bold}.}
\label{tab:v2w}
\centering
\footnotesize
\setlength{\tabcolsep}{8pt}
\begin{tabular}{@{}lcccc@{}}
\toprule
Model & FS & VS & Avg & DSR \\
\midrule
\multicolumn{5}{@{}l}{\textit{Agent (+ Claude Code)}} \\
Claude-Opus-4.8~\cite{anthropic2026opus48} & 65.9 & \underline{59.1} & \underline{62.5} & 98.5 \\
GPT-5.4~\cite{openai2026gpt54} & 55.2 & 30.8 & 43.0 & 98.5 \\
Gemini-3-Pro~\cite{google2025gemini3pro} & 14.6 & 13.3 & 14.0 & 65.2 \\
Kimi-K2.5~\cite{team2026kimi} & 61.0 & 35.1 & 48.1 & 92.4 \\
Step-3.7-Flash~\cite{stepfun2026step37flash} & 47.3 & 22.1 & 34.7 & 95.5 \\
Seed-1.8~\cite{seed2026seed1} & 13.3 & 7.0 & 10.2 & 36.4 \\
Qwen3.5-397B-A17B~\cite{qwen3.5} & \underline{67.1} & 35.6 & 51.4 & 100.0 \\
Qwen3.8-27B~\cite{qwen38} & 45.7 & 36.4 & 41.1 & 48.5 \\
Qwen3-VL-32B-Instruct~\cite{bai2025qwen3} & 0.0 & 0.0 & 0.0 & 0.0 \\
\midrule
\multicolumn{5}{@{}l}{\textit{One-shot}} \\
MiMo-V2.6-Flash~\cite{qiao2026mimo} & 54.3 & 45.0 & 	
49.7 & 100.0 \\
Qwen3.8-Flash-Next~\cite{qwen3.8flashnext} & 51.1 & 41.9 & 46.5 & 100.0 \\
Qwen3.8-27B~\cite{qwen38} & 45.3 & 36.3 & 40.8 & 92.4 \\
Qwen3.5-9B~\cite{qwen3.5} & 20.2 & 22.2 & 21.2 & 95.5 \\
Qwen3.5-4B~\cite{qwen3.5} & 12.9 & 14.1 & 13.5 & 97.0 \\
\rowcolor{cvprblue!8}
\textbf{\icoder{}-4B} & \textbf{51.1} & \textbf{35.5} & \textbf{43.3} & \textbf{93.9} \\
\rowcolor{cvprblue!8}
\quad \textit{vs.\ Qwen3.5-4B (Baseline)} & \textcolor{Seattle}{\textbf{+38.2}} & \textcolor{Seattle}{\textbf{+21.4}} & \textcolor{Seattle}{\textbf{+29.8}} & --- \\
\rowcolor{cvprblue!8}
\textbf{\icoder{}-9B} & \textbf{55.2} & \textbf{41.3} & \textbf{48.2} & \textbf{95.5} \\
\rowcolor{cvprblue!8}
\quad \textit{vs.\ Qwen3.5-9B (Baseline)} & \textcolor{Seattle}{\textbf{+35.0}} & \textcolor{Seattle}{\textbf{+19.1}} & \textcolor{Seattle}{\textbf{+27.0}} & --- \\
\bottomrule
\end{tabular}
\vspace{-4mm}
\end{table}

\subsection{Main Results}
\label{sec:main-results}

\begin{table}[t]
\caption{\textbf{Main results on \arena{}} (one-shot). The strongest baseline is \underline{underlined}; our models are in \textbf{bold}.}
\label{tab:main}
\centering
\footnotesize
\setlength{\tabcolsep}{8pt}
\begin{tabular}{@{}lcccc@{}}
\toprule
Model & Gate & $F$ & $V$ & Score \\
\midrule
GLM-5.3-Flash~\cite{zeng2026glm} & 92.5 & \underline{47.40} & 50.83 & \underline{49.11} \\
Qwen3.8-Flash-Next~\cite{qwen3.8flashnext} & 90.5 & 41.97 & \underline{53.94} & 47.95 \\
Gemini-3.1-Pro-Preview~\cite{gemini31propreview2026} & 97.5 & 35.41 & 45.93 & 40.67 \\
DeepSeek-V4.1-Flash~\cite{deepseekai2026deepseekv41flash} & 96.0 & 38.10 & 42.66 & 40.40 \\
GPT-6 Luna~\cite{openai2026gpt6luna} & 85.0 & 32.18 & 42.09 & 37.14 \\
GPT-5.6 Luna~\cite{openai2026gpt56} & 92.0 & 22.43 & 32.92 & 27.68 \\
Qwen3.8-27B~\cite{qwen38} & 61.0 & 27.97 & 34.05 & 31.01 \\
Qwen3.5-9B~\cite{qwen3.5} & 53.5 & 8.78 & 12.77 & 10.77 \\
Qwen3.5-4B~\cite{qwen3.5} & 47.5 & 7.42 & 10.41 & 8.92 \\
\rowcolor{cvprblue!8}
\textbf{\icoder{}-4B} & \textbf{95.5} & \textbf{67.26} & \textbf{78.53} & \textbf{72.89} \\
\rowcolor{cvprblue!8}
\quad \textit{vs.\ Qwen3.5-4B (Baseline)} & \textcolor{Seattle}{\textbf{+48.0}} & \textcolor{Seattle}{\textbf{+59.84}} & \textcolor{Seattle}{\textbf{+68.12}} & \textcolor{Seattle}{\textbf{+63.97}} \\
\rowcolor{cvprblue!8}
\textbf{\icoder{}-9B} & \textbf{97.0} & \textbf{72.00} & \textbf{82.75} & \textbf{77.38} \\
\rowcolor{cvprblue!8}
\quad \textit{vs.\ Qwen3.5-9B (Baseline)} & \textcolor{Seattle}{\textbf{+43.5}} & \textcolor{Seattle}{\textbf{+63.22}} & \textcolor{Seattle}{\textbf{+69.98}} & \textcolor{Seattle}{\textbf{+66.61}} \\
\bottomrule
\end{tabular}
\vspace{-3mm}
\end{table}

\paragraph{Vision2Web.}
\icoder{} attains the highest functional score among one-shot models and matches multi-turn coding agents on Vision2Web L2 (Table~\ref{tab:v2w}). \icoder{}-9B achieves nearly 3$\times$ the functional score of its base model (20.2 to \textbf{55.2} FS, +35.0) and over 2$\times$ its overall score (21.2 to \textbf{48.2} Avg), surpassing larger backbones such as Qwen3.8-Flash-Next (46.5 Avg) and Qwen3.8-27B (40.8 Avg). \icoder{}-4B reaches \textbf{51.1} FS and \textbf{43.3} Avg, matching Qwen3.8-Flash-Next in functional score and outperforming the 27B variant in both one-shot and agentic settings. 
Against multi-turn coding agents, one-shot \icoder{}-9B (48.2 Avg) matches the 1T-parameter Kimi-K2.5 agent (48.1 Avg) and surpasses GPT-5.4 (43.0 Avg) because execution-grounded training internalizes cross-page and in-page interactions.

\paragraph{\arena{}.}
On our in-distribution \arena{}, our 9B and 4B models outperform all frontier and open-weight baselines evaluated by more than \textbf{23} points (Table~\ref{tab:main}). \icoder{}-9B achieves the highest overall score (\textbf{77.38}), lifting its base model by +66.61 points. Compact \icoder{}-4B ranks second overall at \textbf{72.89} (+63.97 over base). Both models surpass the strongest frontier baselines, GLM-5.3-Flash (49.11) and \mbox{Qwen3.8-Flash-Next} (47.95), by +23.78 to +29.43 points, confirming that reverse synthesis and transition-distilled RL jointly resolve multi-page routing and stateful widgets.

\subsection{Ablation Study}
\label{sec:isdpo-results}

\begin{table}[t]
\caption{\textbf{Ablation of training stages and components}.
SFT trains for 2 epochs on \icorpus{}. GRPO and all three SDPO variants train for a matched 100 RL steps from the SFT checkpoint.}
\label{tab:isdpo}
\label{tab:isdpo-components}
\label{tab:qwen35-4b}
\centering
\footnotesize
\setlength{\tabcolsep}{2.5pt}
\begin{tabular}{@{}lcccccc@{}}
\toprule
 & \multicolumn{3}{c}{Vision2Web L2} & \multicolumn{3}{c}{\arena{}} \\
\cmidrule(lr){2-4}\cmidrule(l){5-7}
Method & FS & VS & Avg & $F$ & $V$ & Score \\
\midrule
\multicolumn{7}{@{}l}{\textit{Qwen3.5-4B~\cite{qwen3.5}}} \\
Zero-shot & 12.93 & 14.05 & 13.49 & 7.42 & 10.41 & 8.92 \\
+ SFT & 21.41 & 18.83 & 20.12 & 57.76 & 71.99 & 64.87 \\
\rowcolor{cvprblue!8}
\textbf{\icoder{}-4B} & \textbf{51.11} & \textbf{35.52} & \textbf{43.31} & \textbf{67.26} & \textbf{78.53} & \textbf{72.89} \\
\midrule
\multicolumn{7}{@{}l}{\textit{Qwen3.5-9B~\cite{qwen3.5}}} \\
Zero-shot & 20.20 & 22.19 & 21.20 & 8.78 & 12.77 & 10.77 \\
+ SFT & 34.34 & 21.18 & 27.76 & 62.35 & 77.07 & 69.71 \\
+ GRPO~\cite{shao2024deepseekmath} & 43.39 & 36.15 & 39.77 & 64.17 & 78.97 & 71.57 \\
\hspace{0.7em}+ SDPO w/ text trace $\tau$ & 46.67 & 36.22 & 41.45 & 64.54 & 78.56 & 71.55 \\
\hspace{0.7em}+ SDPO w/ $\tau$ + static view & 49.71 & 37.02 & 43.36 & 63.93 & 78.49 & 71.21 \\
\hspace{0.7em}+ SDPO w/ $\tau$ + montage $M$ & \textbf{50.51} & \textbf{38.61} & \textbf{44.56} & \textbf{66.16} & \textbf{81.31} & \textbf{73.74} \\[1.5pt]
\rowcolor{cvprblue!8}
\textbf{\icoder{}-9B} & \textbf{55.15} & \textbf{41.29} & \textbf{48.22} & \textbf{72.00} & \textbf{82.75} & \textbf{77.38} \\
\bottomrule
\end{tabular}
\vspace{-3mm}
\end{table}

SFT on \icorpus{} and Stage~2 RL play complementary roles across both model scales: SFT establishes multi-page synthesis on \arena{}, whereas execution-guided RL and self-distillation drive the largest gains on out-of-distribution Vision2Web L2 tasks (Table~\ref{tab:isdpo}). 
Two-epoch SFT provides the foundational leap on \arena{} (8.92 to 64.87 on 4B; 10.77 to 69.71 on 9B). 
On Vision2Web L2, however, token-level imitation is prone to repetitive generation loops on long data arrays and repeated DOM blocks, exhausting the decoding budget before closing the document (27.76 Avg on 9B). 
Because Eq.~\eqref{eq:reward} assigns $g(Y)=0$ to truncated outputs, GRPO directly penalizes non-terminating repetition, improving delivery and lifting Vision2Web L2 by +9.05 FS and +14.97 VS (to 39.77 Avg) and \arena{} to 71.57.

The $2\times2$ transition montage $M$ outperforms both a text-only trace and a static post-action view under matched 100 RL steps (Table~\ref{tab:isdpo}). Adding the replay trace $\tau$ alone lifts Vision2Web L2 by +3.28 FS, to 46.67 FS and 41.45 Avg, while \arena{} Score stays level with GRPO (71.55 vs.\ 71.57). 
In place of $M$, one static post-action student screenshot increases Vision2Web L2 further, to 49.71 FS and 43.36 Avg. 
Pairing $\tau$ with the before/after montage $M$ reaches 50.51 FS and 44.56 Avg on Vision2Web L2 (+7.12 FS and +4.79 Avg over GRPO) and 73.74 on \arena{}, where $V$ rises from 78.97 to 81.31. Relative to the static view, the montage adds +2.53 Score and +2.82 $V$ on \arena{}. 
Extending Interactive-SDPO to 200 steps yields \icoder{}-4B / 9B, more than doubling the post-SFT functional score on 4B (+29.70 FS) and adding +20.81 FS on 9B.

\subsection{Further Analysis}
\label{sec:scaling}
\label{sec:data-ablation}

\begin{figure*}[!t]
\begin{minipage}[t]{0.44\textwidth}
\captionsetup{type=table}
\captionof{table}{\textbf{SFT data comparison on \arena{}}.}
\label{tab:scaling}
\centering
\footnotesize
\setlength{\tabcolsep}{7pt}
\begin{tabular}{@{}lcccc@{}}
\toprule
Training data & Gate & $F$ & $V$ & Score \\
\midrule
Zero-shot (Qwen3.5-9B~\cite{qwen3.5}) & 53.5 & 8.78 & 12.77 & 10.77 \\
\midrule
\multicolumn{5}{@{}l}{\textit{Corpus comparison (10K, 1 epoch)}} \\
WebSight~\cite{laurenccon2024unlocking} & 72.0 & 27.35 & 33.92 & 30.64 \\
\webvibe{} (w/o verification) & 76.5 & 40.12 & 54.07 & 47.10 \\
\midrule
\multicolumn{5}{@{}l}{\textit{\icorpus{} scaling}} \\
SFT 5K (1 epoch) & 72.5 & 43.30 & 52.36 & 47.83 \\
SFT 10K (1 epoch) & 82.0 & 48.29 & 60.86 & 54.58 \\
SFT 20K (1 epoch) & 84.5 & 53.26 & 66.50 & 59.88 \\
SFT 32K (1 epoch) & \underline{87.0} & \underline{56.17} & \underline{68.71} & \underline{62.44} \\
\rowcolor{cvprblue!8}
\textbf{SFT 32K (2 epochs)} & \textbf{94.5} & \textbf{62.35} & \textbf{77.07} & \textbf{69.71} \\
\bottomrule
\end{tabular}
\end{minipage}\hfill
\begin{minipage}[t]{0.54\textwidth}
\vspace{0pt}
\centering
\captionsetup{type=figure}
\includegraphics[width=\linewidth]{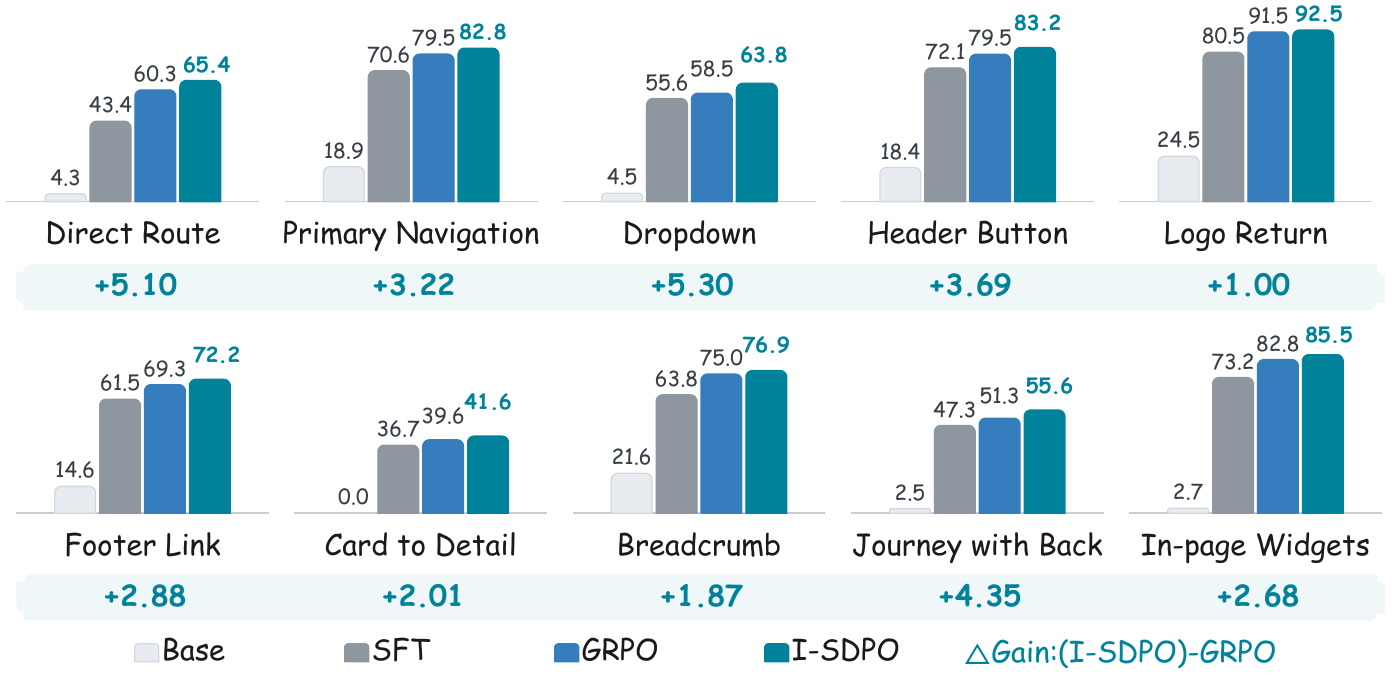}
\vspace{-6mm}
\captionof{figure}{\textbf{Per-category check pass rates on \arena{}}.}
\label{fig:fine-grained}
\end{minipage}

\vspace{4mm}
\centering
\captionsetup{type=figure}
\includegraphics[width=\textwidth]{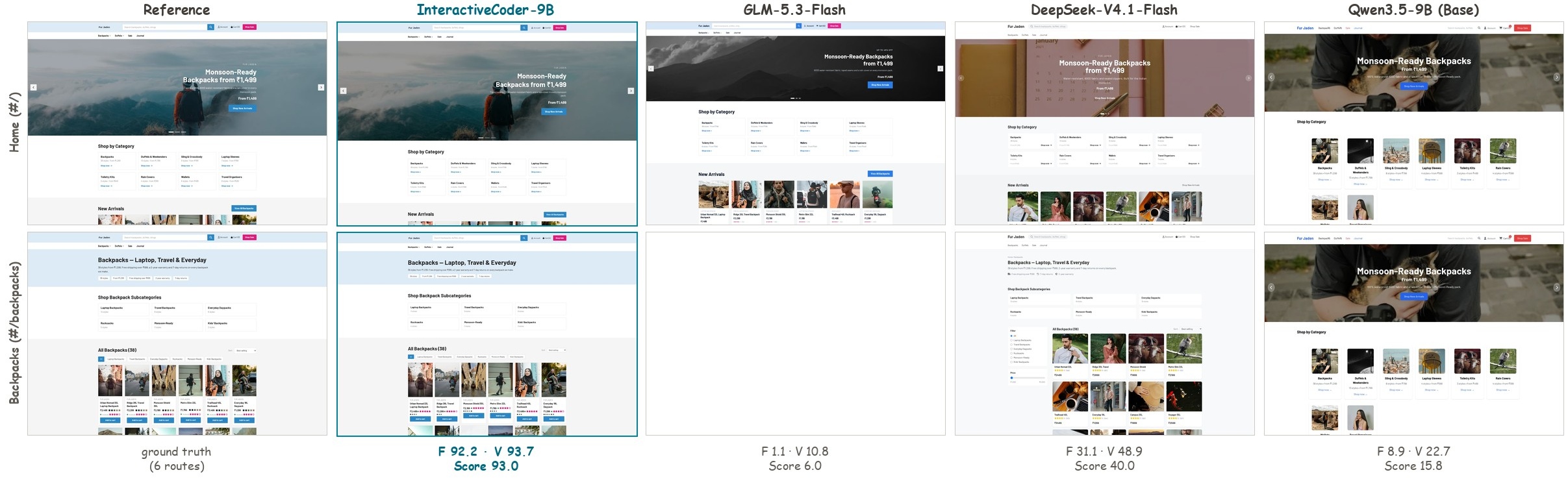}
\vspace{-4mm}
\caption{\textbf{Qualitative comparison on \arena{} site\_009688}. Whole-site $F$, $V$, and Score are reported below each column.}
\label{fig:case-wva}
\vspace{-2mm}
\end{figure*}

\paragraph{Inference Cost.}

One-shot asset grounding from rendered contact sheets removes the multi-turn file-system exploration loop that compact models struggle to execute (Table~\ref{tab:v2w}): Qwen3-VL-32B scores 0.0 under the agentic harness and Qwen3.8-27B drops from 92.4\% one-shot DSR to 48.5\% agentic DSR. 
In a single pass, \icoder{} consumes only about 60K cumulative input-plus-output tokens per project on Vision2Web (Figure~\ref{fig:teaser}d): 
\icoder{}-9B matches the Kimi-K2.5 agent (1T total / 32B activated parameters, 4.80M tokens across 114.3 turns) with 111$\times$ fewer parameters and 78$\times$ fewer tokens, while both 9B and 4B outperform the Step-3.7-Flash (6.05M tokens across 108.8 turns) and Qwen3.8-27B (18.99M tokens across 186.1 turns) agents at 98--310$\times$ lower token cost.

\paragraph{SFT Data Scaling and Source Comparison.}
On \arena{} (Table~\ref{tab:scaling}), 10K static screenshot-to-code pairs from WebSight~\cite{laurenccon2024unlocking} provide no supervision for multi-page routing or stateful widgets (30.64 Score). 
Removing the verified gates from \webvibe{} drops Score from 54.58 to 47.10 ($-8.17$ $F$), confirming the value of execution-gated filtering. 
Scaling verified data from 5K to 32K yields steady gains across all metrics (47.83 to 62.44 at 1 epoch), and another training epoch further lifts Score to \textbf{69.71}.


\vspace{-4mm}
\paragraph{Fine-Grained Capability Analysis.}
\mbox{Interactive-SDPO} consistently improves pass rates, with the largest gains on conditional navigation and multi-step state transitions (Figure~\ref{fig:fine-grained}). 
Based on Qwen3.5-9B, it outperforms GRPO in every category, led by dropdown menus (+5.30), direct routes (+5.10), journeys with Back (+4.35), and header buttons (+3.69), as the transition montage $M$ directly contrasts expected and observed views.

\paragraph{Case Study.}
\label{sec:case-study}
Figure~\ref{fig:case-wva} compares generations on \arena{}. \icoder{}-9B achieves Score $93.0$ ($F=92.2$, $V=93.7$), grounding the exact hero photo and product thumbnails from the contact sheets while wiring category filters and detail routes cleanly on \texttt{\#/backpacks}. In contrast, GLM-5.3-Flash (Score $6.0$) substitutes an unrelated hero image and degrades category tiles into plain text links, DeepSeek-V4.1-Flash (Score $40.0$) shifts the palette and breaks the product grid, and the base model (Score $15.8$) repeats broken banners with non-functional routing.

\section{Conclusion}
\label{sec:conclusion}
\label{sec:limitations}

We presented \webvibe{} and \mbox{Interactive-SDPO} to train \icoder{} for interactive website generation. \webvibe{} reverses synthesis and derives multimodal inputs strictly from browser-verified interactions to build \icorpus{}. During Reinforcement Learning, \mbox{Interactive-SDPO} conditions a weight-sharing self-teacher on the interaction trace and a \mbox{$2\times2$} transition montage, providing token-level supervision at zero inference-time overhead. 
InteractiveCoder attains the highest one-shot functional score on Vision2Web L2, approaches multi-turn agents at far lower token cost, and leads all baselines on the in-distribution WebVibeBench.

\clearpage

\fi

\ifwvmain
\wvprintbibliography
\fi

\ifwvappendix
\ifwvmain
  \clearpage
\else
  \setcounter{page}{1}
\fi
\appendix
\setcounter{table}{0}
\setcounter{figure}{0}
\renewcommand{\thetable}{A.\arabic{table}}
\renewcommand{\thefigure}{A.\arabic{figure}}
\twocolumn[{%
\centering
\Large\bfseries \thetitle\\[5mm]
\Large Supplementary Materials\par
\vspace{2mm}
\begin{center}
\captionsetup{type=figure}
\begin{minipage}[t]{0.49\textwidth}\centering
\includegraphics[width=\linewidth]{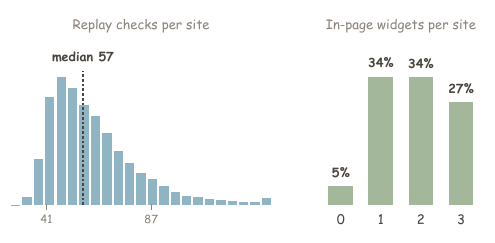}\\[1pt]
{\small (a) Replay checks and in-page widgets per site}
\end{minipage}\hfill
\begin{minipage}[t]{0.49\textwidth}\centering
\includegraphics[width=\linewidth]{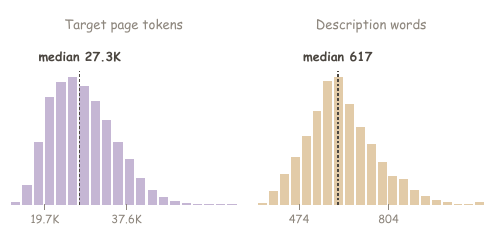}\\[1pt]
{\small (b) Target page and description length}
\end{minipage}
\vspace{-2mm}
\captionof{figure}{\textbf{Interaction density and sequence length distributions of \icorpus{}.}
(a) Browser replay checks passed per site, covering routes, navigation, journeys with Back, and widgets, alongside the number of verified in-page widgets per site. (b) Target page tokens and description words. Dashed lines mark medians and grey ticks the 10th and 90th percentiles; the mean full sequence is 37{,}974 tokens.}
\label{fig:corpus-stats-extra}
\end{center}
}]

\hyperref[app:dp]{Appendix~A} details the \webvibe{} reverse-synthesis engine.
\hyperref[app:arena]{Appendix~B} describes the benchmarks (Vision2Web and WebVibeBench) and the shared evaluation protocol.
\hyperref[app:training]{Appendix~C} records the training configuration for both model scales and shows how interactive visual feedback acts as privileged supervision.
\hyperref[app:more-cases]{Appendix~D} presents additional qualitative comparisons on \arena{} and Vision2Web L2, and \hyperref[app:prompts]{Appendix~E} reproduces the prompts used in synthesis and evaluation.

\section{WebVibe Construction Details}
\label{app:dp}

This section specifies the \webvibe{} reverse-synthesis engine summarized in Section~3.
Table~\ref{tab:dp-stages} lists the stages of LLM and MLLM in the pipeline along with their backbone models and functions, and Figure~\ref{fig:corpus-stats-extra} reports the resulting distributions in the 31{,}542 accepted sites in \icorpus{}.

\subsection{Task Formulation and Contract Compilation}
\label{app:dp-format}
\label{sec:dp-seed}
\label{sec:dp-plan}
\label{app:dp-compile}

\paragraph{Sample Format.}
Each sample $(X,Y)$ follows Eq.~(1) under the unified instruction template of Prompt~\ref{prompt:1}, which specifies the runtime, hash-routing, layout, and asset-grounding rules shared across training and evaluation. Images are ordered with route screenshots first (home page followed by remaining pages in visitor reachability order, captured at a 1920\,px viewport and capped at 8000\,px in height) and visual asset contact sheets second. Each contact sheet packs up to $8\times8$ labeled square tiles, with font files rendered as specimen tiles in their own typeface. The target $Y$ is the verified single-file application after rewriting its asset paths to the anonymized filenames listed in $X$.

\paragraph{Style Cards.}
To prevent the template homogeneity typical of synthetic webpages, every site is conditioned on an offline \emph{style card} (Prompt~\ref{prompt:2}) extracted once by an MLLM from deduplicated real-webpage screenshots and merged into site-level appearance descriptors using a fixed categorical vocabulary. Style cards capture only visual layout attributes, such as color schemes, page structures, geometric features, and density. A synthetic site thus inherits a realistic visual style and page-height distribution without reusing any source text, imagery, or code.

\paragraph{Site Planning and Contract Compilation.}
Conditioned on an organization metadata seed and the style card, the LLM planner (Prompt~\ref{prompt:4}) synthesizes a structured JSON plan of pages, sections, data collections with detail views, navigation links, and in-page widgets. A deterministic compiler validates structural reachability and lowers the plan into resolved hash routes (\texttt{\#/}$p$, \texttt{\#/}$c$\texttt{/}$k$), typed image and logo slots, the prototype capture schedule, and the executable verification contract $\mathcal{C}_X$ covering all routes, navigation links, multi-step journeys with Back retracing, and in-page widgets. The code generator receives only the layout plan, style card, and asset manifest, and never sees $\mathcal{C}_X$.

\subsection{Synthesis and Replay Verification}
\label{sec:dp-assets}
\label{sec:dp-page}
\label{app:dp-verify}

\paragraph{Asset Grounding.}
All visual assets are selected and frozen prior to code synthesis so that the application is written around concrete visual files. Candidate photographs are retrieved from a local index of 10M Pexels images annotated with an 18{,}440-tag vocabulary via exact tag match, multi-word tag intersection, head-noun fallback, and aspect-ratio filtering (portrait for profiles, square for products and testimonials, and landscape for heroes and banners), without reusing any image within a site. Because automated tags often describe background elements rather than foreground subjects, an MLLM auditor (Prompt~\ref{prompt:3}) inspects candidate picks against their target slot descriptions and rejects off-subject or text-dominated images (slides, advertisements, document scans, screenshots, charts, or collages) for up to three replacement rounds; any unresolved slot discards the site. Interface icons are retrieved from Iconify Material Design, brand logos are generated as SVG wordmarks, and open-licensed webfont files are drawn from a 70-family pool across eight typographic classes under real-world filename conventions (\texttt{Poppins-Bold.woff2}, Fontsource slugs, Google-static hashes, or short hex strings) so typefaces must be identified visually from their specimen tiles.

\paragraph{Page Synthesis.}
A code LLM (Prompt~\ref{prompt:5}) synthesizes the self-contained single-file application from the compiled plan, style card, frozen asset manifest, and assigned fonts. Generation is prefilled with \texttt{<!DOCTYPE html>} and capped at 65{,}536 tokens.

\paragraph{Static Verification.}
Before launching a browser, a static verifier enforces three pre-execution gates: (i)~\emph{offline self-containment}, requiring the local React/Tailwind/Babel vendor bundle and rejecting any external URL, network call (\texttt{fetch}, \texttt{import}, \texttt{require}), non-deterministic API (\texttt{Math.random}), lazy loading, or string-concatenated JSX asset path; (ii)~\emph{asset and typography binding}, requiring every declared font face and at least 70\% of photo slots to be referenced with no unresolved slot tokens; and (iii)~\emph{anti-degeneracy}, requiring structured data rendering ($\ge 3$ array \texttt{map} calls, file size $\ge 18$\,KB, and student-tokenized length within budget) with no line repeated more than five times and no placeholder copy (e.g., ``lorem ipsum'').

\begin{table}[t]
\caption{\textbf{LLM and MLLM stages in the \webvibe{} synthesis.}}
\label{tab:dp-stages}
\centering
\footnotesize
\setlength{\tabcolsep}{3.5pt}
\begin{tabularx}{\linewidth}{@{}l l Y@{}}
\toprule
Stage & Model & Role \\
\midrule
Style card (\hyperref[prompt:2]{P\ref*{prompt:2}}) & Qwen3.8-Flash-Next & Extract visual style \\
Site plan (\hyperref[prompt:4]{P\ref*{prompt:4}}) & DeepSeek-V4.1-Flash & Synthesize JSON site plan \\
Asset audit (\hyperref[prompt:3]{P\ref*{prompt:3}}) & Qwen3.8-Flash-Next & Filter candidate photos \\
Page synthesis (\hyperref[prompt:5]{P\ref*{prompt:5}}) & DeepSeek-V4.1-Flash & Generate webpage \\
Code repair (\hyperref[prompt:6]{P\ref*{prompt:6}}) & GLM-5.3 & Patch failed checks via diffs \\
Visual judge (\hyperref[prompt:7]{P\ref*{prompt:7}}) & Qwen3.8-Flash-Next & Audit route screenshots \\
Description (\hyperref[prompt:8]{P\ref*{prompt:8}}) & Gemini-3.1-Pro & Write task description \\
\bottomrule
\end{tabularx}
\end{table}

\paragraph{Browser Replay.}
Accepted candidates are served statically in headless Chromium and replayed against $\mathcal{C}_X$, locating controls deterministically by accessible text within their landmark region (\texttt{header}, \texttt{main}, \texttt{footer}). A route is verified when \texttt{location.hash} and the primary \texttt{<h1>} match the compiled plan with a non-empty mounted view, and multi-step journeys must retrace cleanly under browser \texttt{Back}. For in-page widgets (Eq.~(2)), let $\mathcal{T}_a,\mathcal{T}_b$ denote the main-content DOM text and $I_a,I_b$ the control-masked viewport captures before and after action $c$: the visibility gate is $\Delta_{\mathrm{vis}}(a,b)=\mathbb{1}[\mathcal{T}_a\neq \mathcal{T}_b]\vee\mathbb{1}[d_{\text{pix}}(I_a,I_b)>\tau_{\text{pix}}]$, while $\Phi_{\mathrm{sem}}(c;\,a\to b)$ requires target item titles to appear and excluded titles to vanish ($\mathcal{I}^+_c\subseteq\mathcal{T}_b \wedge \mathcal{I}^-_c\cap\mathcal{T}_b=\emptyset$) for filters and search, target panel or confirmation snippets to flip ($s^\star_c\notin\mathcal{T}_a \wedge s^\star_c\in\mathcal{T}_b$) for tabs, accordions, and forms, and masked slide transitions to register in $\mathcal{P}_X$ for carousels. Candidates failing static, replay, or render checks undergo at most two search/replace repair turns (Prompt~\ref{prompt:6}) before being discarded.

\subsection{Reverse Input Derivation and Packaging}
\label{sec:dp-judge}
\label{sec:dp-desc}
\label{app:dp-pack}

\paragraph{Visual Auditing.}
An MLLM visual auditor (Prompt~\ref{prompt:7}) inspects all route captures of a candidate site and rejects sites exhibiting broken layouts, blank viewports, blocking overlays, poster-like non-webpages, off-topic imagery, or cross-page visual inconsistency. 

\paragraph{Grounded Task Description.}
A text-only specification writer LLM (Prompt~\ref{prompt:8}) synthesizes the first-person owner-perspective task description ($\text{Text}$) strictly from verified DOM facts and replay-confirmed routes, journeys, and widgets. Automated post-checks enforce natural prose and exact substring grounding of every quoted UI label in the verified DOM or replay log.

\paragraph{Distractor Injection and Contact-Sheet Packaging.}
To prevent filename shortcuts and mirror real-world frontend repositories where used assets coexist with unused project files, the packager injects distractor assets before rendering the contact sheets. Used assets and distractors are shuffled together, anonymized to sequential filenames per modality (\texttt{img\_7.jpg}, \texttt{icon\_3.svg}, \texttt{logo\_1.svg}), and packed onto at most ten $8\times8$ contact sheets. Each released sample stores only the student-facing pair $(X,Y)$ while exporting $(\mathcal{C}_X,\mathcal{P}_X)$ for Stage~2 RL.

\section{Benchmarks}
\label{app:arena}
\label{app:modes}
\label{app:bench-adapters}

We evaluate all models on two interactive frontend benchmarks under a unified protocol. Every one-shot model receives the same instruction template (Prompt~\ref{prompt:1}), textual specification, route prototypes, and visual asset contact sheets. Decoding uses temperature 0.6, top-$p$ 0.95, and top-$k$ 20 with a 65{,}536-token budget (with presence penalty 1.5 and repetition penalty 1.0). All one-shot evaluations strictly grade the first generation pass per task without retries; launch failures and truncated outputs remain in the denominator and receive a score of 0. 

\subsection{Vision2Web}
\label{app:eval-v2w}

Vision2Web~\cite{he2026vision2web} is a benchmark for multimodal website development spanning static webpages, interactive frontends, and full-stack applications. We evaluate on the Interactive Frontend level (L2), which comprises 66 multi-page frontend projects. Each task provides UI prototype screenshots, a requirement document, and project assets, requiring a runnable application with visual fidelity and coherent cross-page and in-page interactions.

For one-shot generation, we pass the requirement document as $\text{Text}$, the prototype screenshots at native resolution as $\text{Visual}$, and the project assets rendered onto $8\times8$ contact sheets as $\text{Asset}$. For the multi-turn \emph{Agent (+ Claude Code)} baselines, each backbone runs through the official Vision2Web \texttt{claude\_code} harness inside an isolated \texttt{vision2web-sandbox} container with bash and file-editing tools until it delivers a runnable application or reaches the task timeout. During evaluation, each delivered project is launched in a fresh $1920\times1080$ browser session, where the official Vision2Web evaluation engine interacts with the running application along the predefined workflows in \texttt{workflow.json} via \texttt{playwright-cli} (executing \texttt{snapshot}, \texttt{click}, \texttt{fill}, and \texttt{go-back} actions over accessibility-tree element references) and captures full-page screenshots at the designated workflow states. 

We report Functional Score (FS), Visual Score (VS), their arithmetic mean (Avg), and Delivery Success Rate (DSR). FS measures the proportion of functional workflow checks passed, and VS aggregates block-wise visual similarity across reference prototypes, both on a 0--100 scale. DSR is the percentage of the 66 projects that successfully launch and render; undelivered projects are retained in the denominator and contribute 0 to both FS and VS.

\subsection{\arena{}}
\label{app:arena-split}
\label{app:arena-metrics}

\arena{} evaluates one-shot multimodal generation of multi-page interactive websites with deterministic browser-replay contracts. The benchmark comprises 200 held-out websites across 87 domains, strictly disjoint from the \icorpus{} SFT and RL splits: every task uses an independent organization seed, site plan, copy, and per-site shuffled and anonymized asset pool (\texttt{img\_$k$.jpg}), with style cards specifying only abstract layout attributes rather than HTML/JSX templates.

During evaluation, each generated single-file application is served statically in a fresh headless browser session. First, the render gate $g(Y)\in\{0,1\}$ checks that the application mounts a non-empty DOM with zero syntax, runtime, or missing-asset errors and renders all local images. Once the gate passes, the deterministic replay harness executes every planned check in the site's compiled contract $\mathcal{C}_X$ without relying on internal test IDs, hidden route slugs, or DOM templates: it resolves controls solely by human-visible labels within semantic landmarks and verifies direct rendering of all top-level and item-detail routes, eight cross-page navigation families (including multi-step journeys with browser \texttt{Back} retracing), and six stateful in-page widget families.

We report the render gate rate (Gate), Functional Score ($F$), Visual Score ($V$), and overall Score (Eq.~(4)). $F$ is the pass rate over all planned checks in $\mathcal{C}_X$, where each planned route, navigation interaction, and in-page widget contributes one unit (with per-category pass rates reported in Figure~5 and form filling plus submission counted as a single check; checks absent from a site plan are excluded from $F$). $V$ averages visual similarity over all planned views, and Score is the gated mean $g(Y)\cdot\frac{1}{2}(F+V)$, assigning 0 to any candidate that fails the render gate.

\section{Training Details}
\label{app:training}
\label{app:rl-details}

Table~\ref{tab:train-config} lists the shared training hyperparameters for both Qwen3.5-4B and Qwen3.5-9B in Section~5. 

\begin{table}[t]
\caption{\textbf{Training settings} for InteractiveCoder.}
\label{tab:train-config}
\centering
\footnotesize
\begin{tabularx}{0.9\linewidth}{@{}Xc@{}}
\toprule
Setting & Qwen3.5-4B / 9B \\
\midrule
\multicolumn{2}{@{}l}{\textit{Supervised fine-tuning}} \\
Framework & ms-swift~\cite{zhao2024swiftascalablelightweightinfrastructure} \\
Samples / epochs & 31{,}542 / 2 \\
Global batch / steps & 32 / 1{,}972\\
Sequence limit (tokens) & 86{,}016 \\
Image pixel bounds & $65{,}536$--$921{,}600$ \\
Learning rate & $2{\times}10^{-5}$ \\
Schedule / warmup & cosine / 3\% \\
Weight decay & 0.1 \\
Grad clip & 1.0 \\
Sequence parallel & 2 \\
Precision & bf16 \\
\midrule
\multicolumn{2}{@{}l}{\textit{Reinforcement learning}} \\
Framework & veRL~\cite{sheng2024hybridflow} \\
Prompts / seed & 2{,}000 / 42 \\
Prompts / step $\times$ rollouts & 8 $\times$ 8 \\
Learning rate & $1{\times}10^{-6}$ (constant) \\
Temperature / top-$p$ & 1.0 / 0.85 \\
Prompt / response cap & 20{,}480 / 50{,}000 \\
GRPO clip $\epsilon$ / grad clip & 0.2 / 1.0 \\
Policy epochs / ref KL & 1 / none \\
SDPO $K$ / weight $\lambda$ / IS clip & 4 / 0.1 / 2.0 \\
Visual reward judge & Qwen3.8-27B \\
Browser workers / timeout & 12$\times$ Chromium 1243 / 150\,s \\
Vision tower / parallelism & frozen / ZeRO-3 \\
\bottomrule
\end{tabularx}
\vspace{-3mm}
\end{table}

\paragraph{Reward Metrics.}
A pool of browser workers scores rollouts: each candidate is written into a fresh copy of its site directory, loaded at 1920\,px, and replayed with the compiled contract. Unlike Stage~1 synthesis (which enforces exact \texttt{location.hash} and \texttt{<h1>} matches), RL replay navigates via visible link labels (falling back to clicking the matching navigation link whenever a direct hash probe does not match the target heading) and verifies reached pages by heading similarity and page-specific text anchors rather than requiring identical hash strings. Infrastructure errors are tracked separately and re-queued rather than scored as 0; the visual reward judge $v_{\text{judge}}$(Qwen3.8-27B) scores only reached states against their reference captures. 

\paragraph{Interactive-SDPO.}
Interactive-SDPO selects at most $K=4$ failed rollouts per group with $\lambda=0.1$ and conditions the weight-sharing self-teacher on interactive visual feedback $h_i=(\tau_i,M_i)$ as privileged supervision. Within each journey, the transition is captured at the first failed action $k_i$, and across checks, failed transitions are ordered worst first for the self-teacher. Whereas a scalar reward or static screenshot cannot distinguish a working interactive control from a visually faithful dead mock-up, Figures~\ref{fig:teacher-context}--\ref{fig:teacher-context-form} show how pairing the structured DOM delta trace $\tau_i$ with the $2\times2$ before-and-after transition montage $M_i$ directly pinpoints broken state transitions across in-page search/filtering, render-gate diagnostics, cross-page route navigation, and stateful form submission.

\begin{figure*}[t]
\centering
\setlength{\fboxsep}{8pt}
\fcolorbox{cvprblue!55!black}{cvprblue!3}{%
\begin{minipage}{0.96\textwidth}
\footnotesize\rmfamily\raggedright
\textbf{\textcolor{cvprblue!80!black}{(a) In-Page Search/Filter Failure ($g(Y_i)=1$): Privileged Trace $\tau_i$ and $2\times2$ State-Transition Montage $M_i$}}\\[3pt]
{\rmfamily
\textcolor{black!65}{Privileged interaction trace. The student did not see this.}\\
\textcolor{black!65}{For each step: what the reference page changed, then what the student's page changed.}\\[3pt]
\textbf{\textcolor{red!70!black}{Step 10: Search for atlas (failed)}}\\
\textbf{failure:} action failed: could not find the field \textbf{``Search parcels\ldots''} in the page content\\
\textbf{\textcolor{Seattle}{expected added:}} \textbf{3 shipments}\\
\textbf{\textcolor{red!70!black}{observed added:}} \textbf{(nothing)}\\
\textbf{\textcolor{Seattle}{expected removed:}} \textbf{8 shipments; River Collection; East Wharf $\cdot$ Tomorrow 11:15}\\
\textbf{\textcolor{red!70!black}{observed removed:}} \textbf{(nothing)}\\[3pt]
\textcolor{black!65}{Transition images follow, one per step, worst first.}\\
\textbf{Image:} Search for atlas\\
\textbf{Transition image 1:} top teacher before/after, bottom student before/after.
}\\[6pt]
\centering
\includegraphics[width=\linewidth]{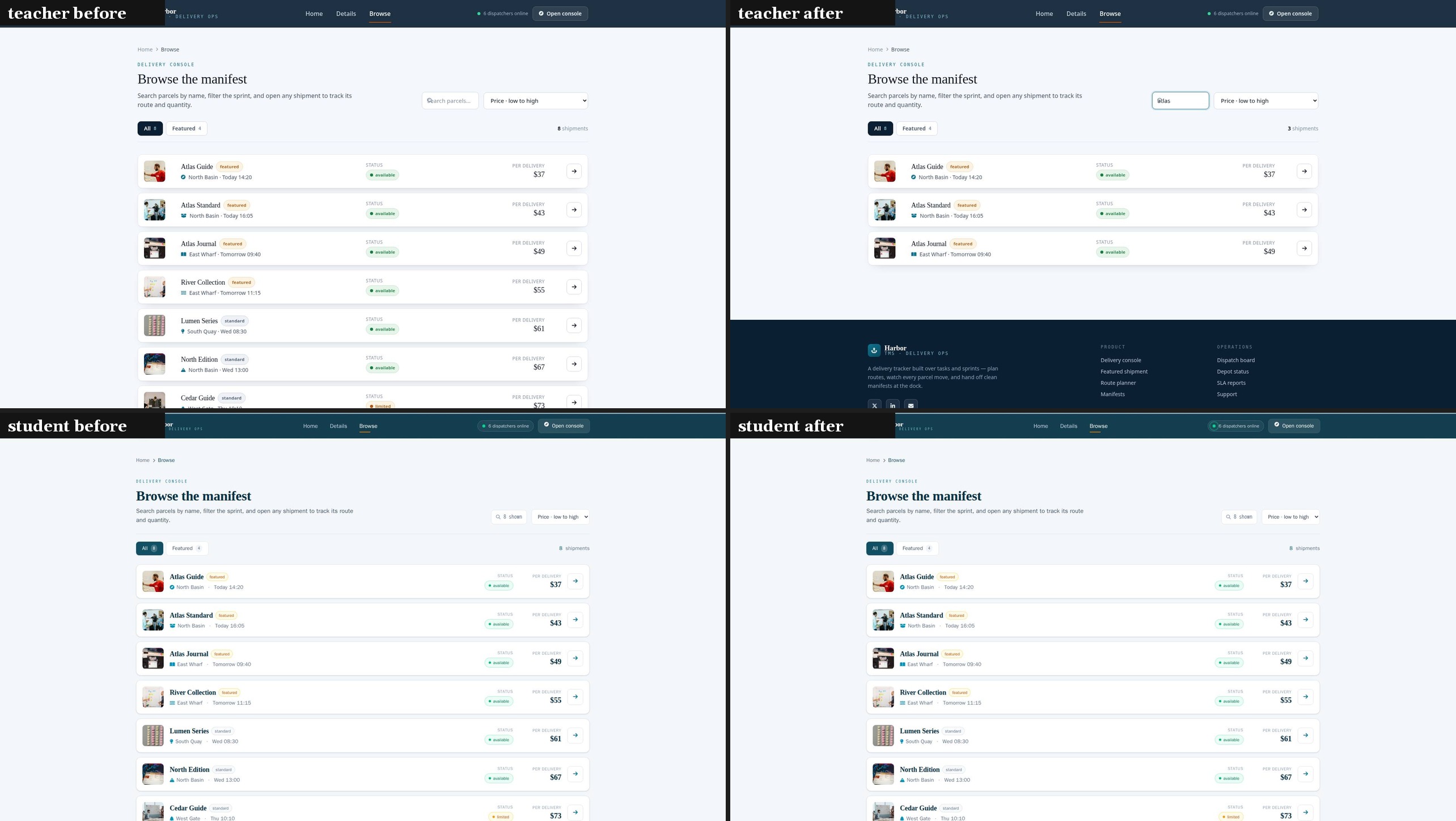}\\[3pt]
{\scriptsize\rmfamily\textcolor{black!75}{\textbf{Top row (Teacher Before $\to$ After):} searching ``atlas'' filters the 8-row manifest down to \textbf{3 matching shipments} and reveals the dark navy footer. \textbf{Bottom row (Student Before $\to$ After):} static appearance matches the reference ($V=0.89$), but the list \textbf{remains unchanged at all 8 rows after the action}.}}\\[6pt]
\raggedright\rule{\textwidth}{0.4pt}\\[5pt]
\textbf{\textcolor{cvprblue!80!black}{(b) Render-Gate Failure ($g(Y_i)=0$): Sanitized Runtime Diagnostic}}\\[3pt]
{\rmfamily
\textbf{\textcolor{red!70!black}{The student's page did not render.}}\\
-- the page renders \textbf{blank or almost empty (0 visible characters)}\\
-- \textbf{first runtime error:} /Inline Babel script: \textbf{Unexpected token, expected ``\}'' (800:227)}\\[2pt]
\hspace*{0.8em}798 $|$ $<$div className="mt-6 bg-paper2 border border-line rounded-2xl p-5"$>$\\
\hspace*{0.8em}799 $|$ \hspace*{1.0em}$<$div className="font-bold text-sm mb-2"$>$Pro tip$<$/div$>$\\
\textbf{\textcolor{red!70!black}{$>$ 800 $|$}} \hspace*{1.0em}$<$p className="text-inksoft \ldots
}
\end{minipage}}
\caption{\textbf{Interactive visual feedback as privileged supervision $h_i$ in Interactive-SDPO (Part 1: In-page filtering and render-gate failure).} (a)~Although the student rollout achieves high static visual fidelity ($V=0.89$), the expected search field is missing and the shipment list remains unfiltered after the action; pairing the DOM delta trace $\tau_i$ with the $2\times2$ transition montage $M_i$ directly contrasts the reference's 8-to-3 filtered transition against the student's unchanged 8-row view. (b)~When a rollout fails the render gate ($g(Y_i)=0$), $h_i$ provides the localized Babel syntax diagnostic and offending source line.}
\label{fig:teacher-context}
\end{figure*}

\begin{figure*}[t]
\centering
\setlength{\fboxsep}{8pt}
\fcolorbox{cvprblue!55!black}{cvprblue!3}{%
\begin{minipage}{0.96\textwidth}
\footnotesize\rmfamily\raggedright
\textbf{\textcolor{cvprblue!80!black}{Cross-Page Route Navigation Failure ($g(Y_i)=1$): Privileged Trace $\tau_i$ and $2\times2$ State-Transition Montage $M_i$}}\\[3pt]
{\rmfamily
\textcolor{black!65}{Privileged interaction trace. The student did not see this.}\\
\textcolor{black!65}{For each step: what the reference page changed, then what the student's page changed.}\\[3pt]
\textbf{\textcolor{red!70!black}{Step 1: Open the Details page (failed)}}\\
\textbf{failure:} expected to reach the page titled \textbf{``Vertical Standard''}, but the page heading is \textbf{``The vertical marketplace for specialist trade goods.''}\\
\textbf{\textcolor{Seattle}{expected added:}} \textbf{Ships in 2 days; Buyer protected; 14-day returns}\\
\textbf{\textcolor{red!70!black}{observed added:}} \textbf{Atlas; VERTICAL MARKETS; Home}\\
\textbf{\textcolor{Seattle}{expected removed:}} \textbf{Independent sellers $\cdot$ specialist trade goods; The vertical marketplace for specialist trade goods.}\\
\textbf{\textcolor{red!70!black}{observed removed:}} \textbf{(nothing)}\\[3pt]
\textcolor{black!65}{Transition images follow, one per step, worst first.}\\
\textbf{Image:} Open the Details page\\
\textbf{Transition image 1:} top teacher before/after, bottom student before/after.
}\\[6pt]
\centering
\includegraphics[width=\linewidth]{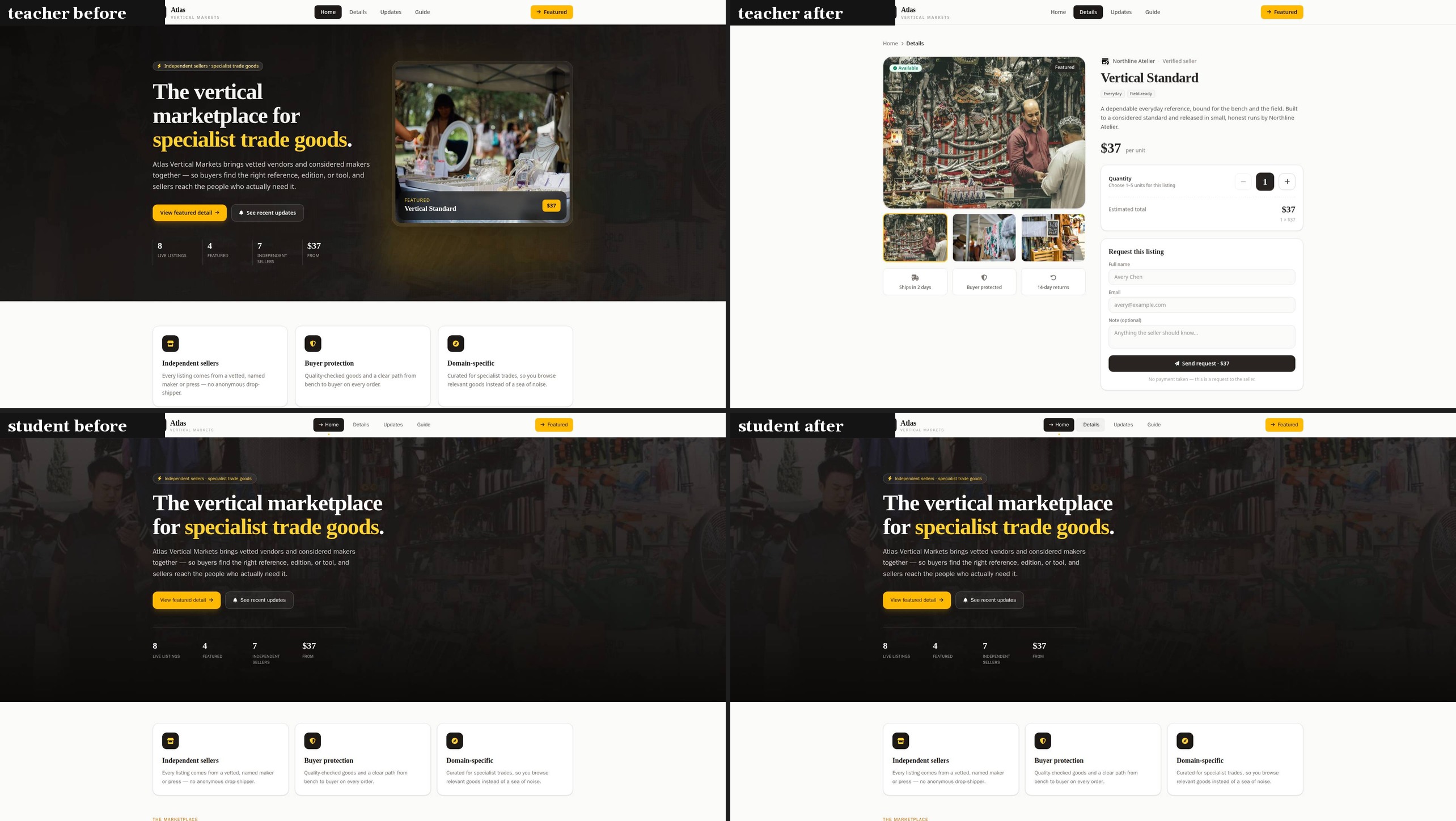}\\[3pt]
{\scriptsize\rmfamily\textcolor{black!75}{\textbf{Top row (Teacher Before $\to$ After):} clicking Details navigates from the dark home hero to the light product detail view (Vertical Standard, \$37) with its gallery and request form. \textbf{Bottom row (Student Before $\to$ After):} the student's click handler fails to switch routes, leaving the dark home view unchanged.}}
\end{minipage}}
\caption{\textbf{Interactive visual feedback as privileged supervision $h_i$ in Interactive-SDPO (Part 2: Cross-page route navigation).} The $2\times2$ transition montage $M_i$ exposes a dead navigation link that static home-page inspection cannot detect: clicking \texttt{Details} switches the reference from the dark landing page to the light item-detail page, whereas the student rollout remains stuck on the home route.}
\label{fig:teacher-context-more}
\end{figure*}

\begin{figure*}[t]
\centering
\setlength{\fboxsep}{8pt}
\fcolorbox{cvprblue!55!black}{cvprblue!3}{%
\begin{minipage}{0.96\textwidth}
\footnotesize\rmfamily\raggedright
\textbf{\textcolor{cvprblue!80!black}{Stateful Form Submission Failure ($g(Y_i)=1$): Privileged Trace $\tau_i$ and $2\times2$ State-Transition Montage $M_i$}}\\[3pt]
{\rmfamily
\textcolor{black!65}{Privileged interaction trace. The student did not see this.}\\
\textcolor{black!65}{For each step: what the reference page changed, then what the student's page changed.}\\[3pt]
\textbf{\textcolor{red!70!black}{Step 16: Submit the local request form (failed)}}\\
\textbf{failure:} expected div \textbf{``Request received for Avery Chen Thanks, Avery Chen. A reply''} in the page content to be visible, but it is not\\
\textbf{\textcolor{Seattle}{expected added:}} \textbf{Request received for Avery Chen; Thanks, Avery Chen. A reply is on its way to avery@example.com.; Reply within 1 business day}\\
\textbf{\textcolor{red!70!black}{observed added:}} \textbf{Sending\ldots}\\
\textbf{\textcolor{Seattle}{expected removed:}} \textbf{Full name; Email address; What can we help with?}\\
\textbf{\textcolor{red!70!black}{observed removed:}} \textbf{Submit request}\\[3pt]
\textcolor{black!65}{Transition images follow, one per step, worst first.}\\
\textbf{Image:} Submit the local request form\\
\textbf{Transition image 1:} top teacher before/after, bottom student before/after.
}\\[6pt]
\centering
\includegraphics[width=\linewidth]{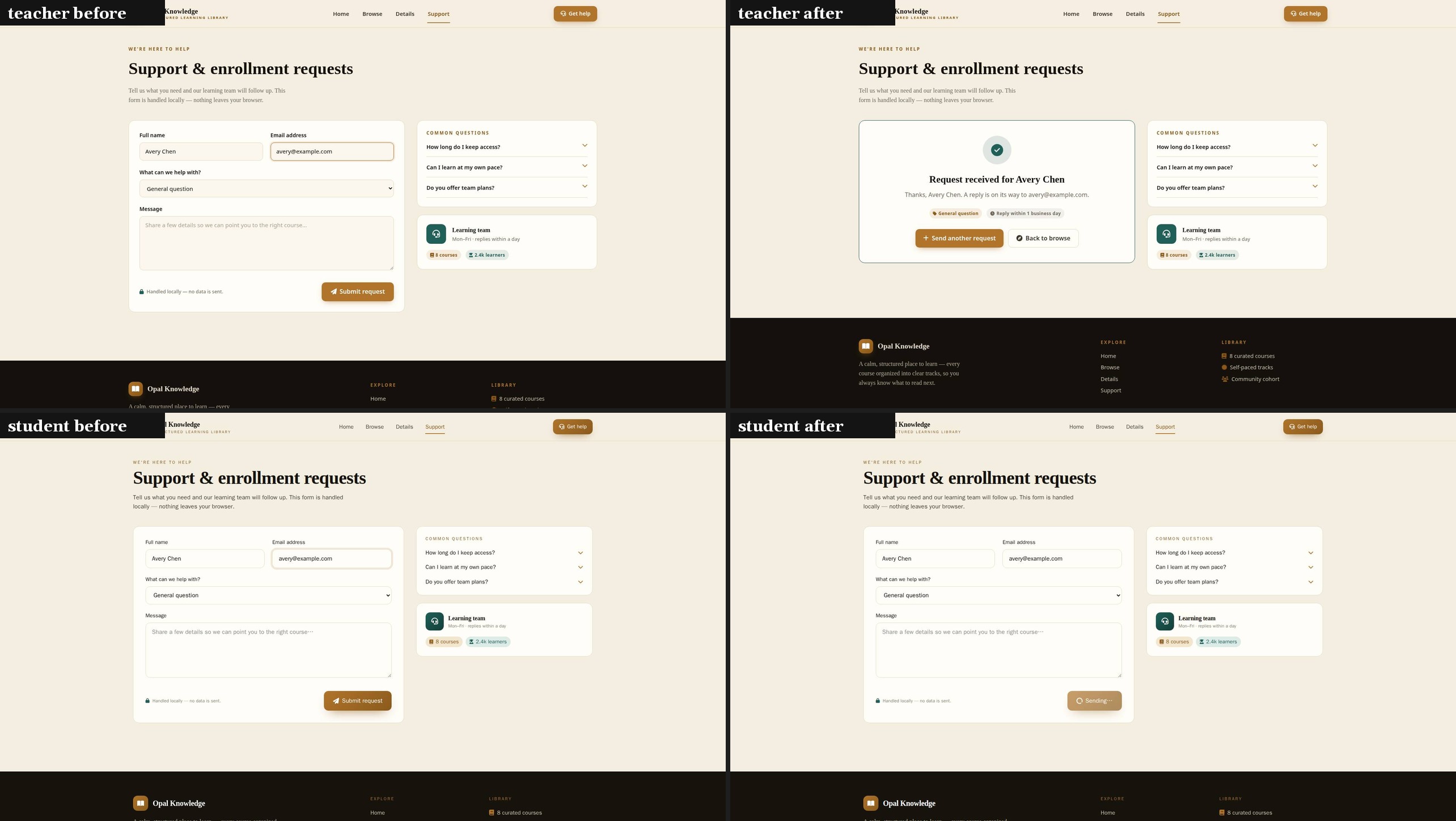}\\[3pt]
{\scriptsize\rmfamily\textcolor{black!75}{\textbf{Top row (Teacher Before $\to$ After):} submitting the filled support form replaces the inputs with a green confirmation card (Request received for Avery Chen). \textbf{Bottom row (Student Before $\to$ After):} the pre-submit form matches the reference ($V=0.93$), but clicking submit leaves the form stuck on a disabled Sending\ldots\ button without rendering the confirmation state.}}
\end{minipage}}
\caption{\textbf{Interactive visual feedback as privileged supervision $h_i$ in Interactive-SDPO (Part 3: Stateful form submission).} Even when the pre-submission form is nearly identical to the reference ($V=0.93$), $h_i=(\tau_i,M_i)$ reveals that the student's submit handler stalls on an intermediate \texttt{Sending\ldots} button instead of transitioning to the confirmation card, providing targeted token-level guidance for the self-teacher.}
\label{fig:teacher-context-form}
\end{figure*}

\section{Additional Case Studies}
\label{app:more-cases}

Figures~\ref{fig:case-wva-app1} and~\ref{fig:case-wva-app2} present eight additional multi-page comparisons on \arena{}, and Figures~\ref{fig:case-v2w-app1} and~\ref{fig:case-v2w-app2} present eight out-of-distribution comparisons on Vision2Web L2, each displaying two routes per project across Ground-Truth Reference, \icoder{}-9B, the strongest one-shot baselines, and the base model. Across both benchmarks, three consistent qualitative patterns emerge: (1)~\emph{exact asset and typography grounding}, where \icoder{}-9B accurately binds the target photographs, icons, and SVG wordmarks from the contact sheets while frontier baselines frequently render broken image placeholders or substitute off-topic assets; (2)~\emph{verified cross-page routing}, where the second column of each pair confirms that \icoder{}-9B switches cleanly to distinct catalog, schedule, or detail views (reaching $88.5$--$98.4$ Score on \arena{} and up to $100.0$ FS on Vision2Web L2), whereas baselines often leave navigation links dead or collapse secondary routes into unstyled lists; and (3)~\emph{out-of-distribution layout generalization}, where training on \icorpus{} and Interactive-SDPO transfers to unseen Vision2Web L2 prototypes and requirement documents without overfitting to synthetic templates.

\begin{figure*}[t]
\centering
\includegraphics[width=\textwidth,height=0.95\textheight,keepaspectratio]{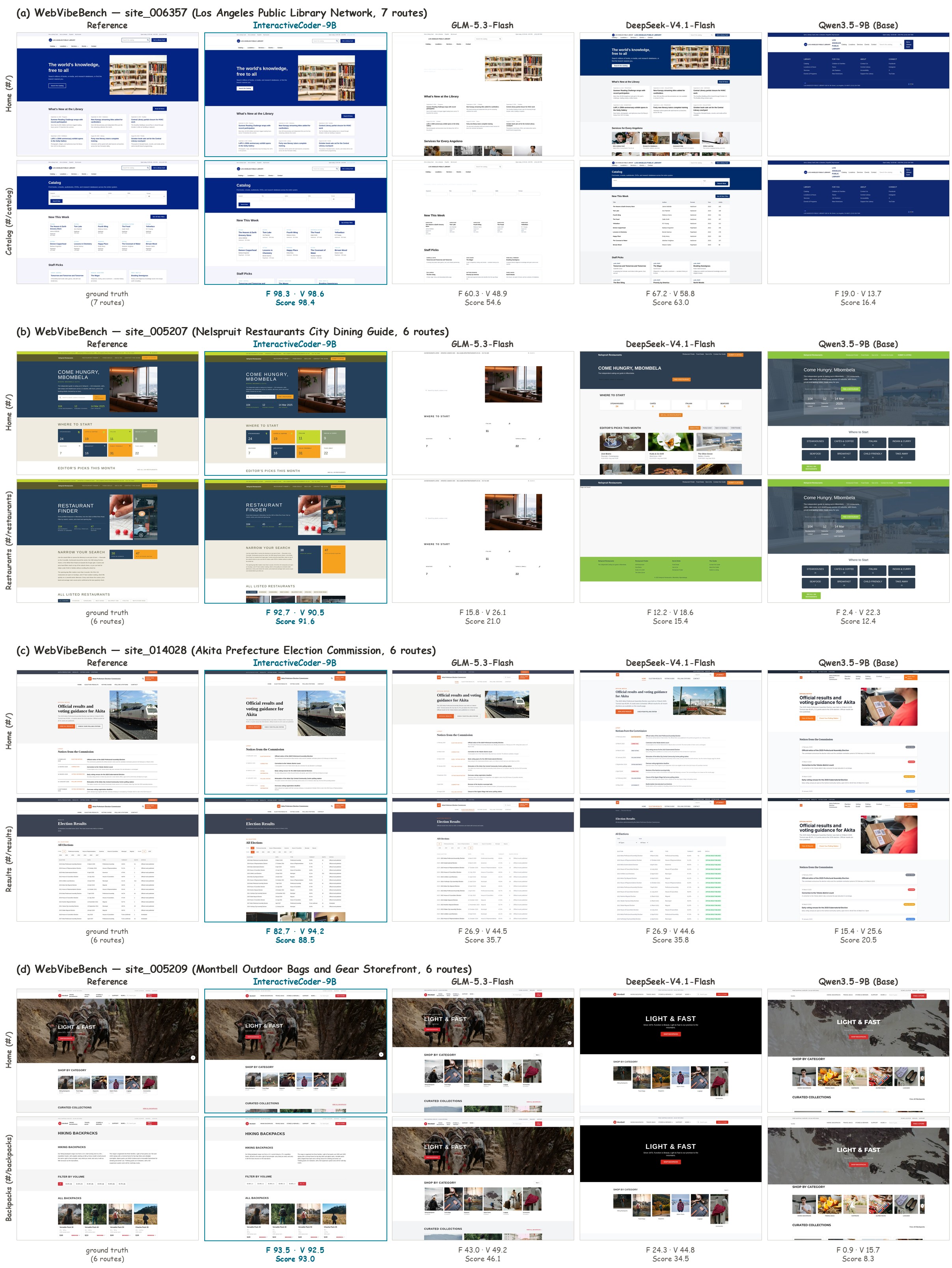}
\caption{\textbf{Additional \arena{} comparisons (1/2).}}
\label{fig:case-wva-app1}
\end{figure*}

\begin{figure*}[t]
\centering
\includegraphics[width=\textwidth,height=0.95\textheight,keepaspectratio]{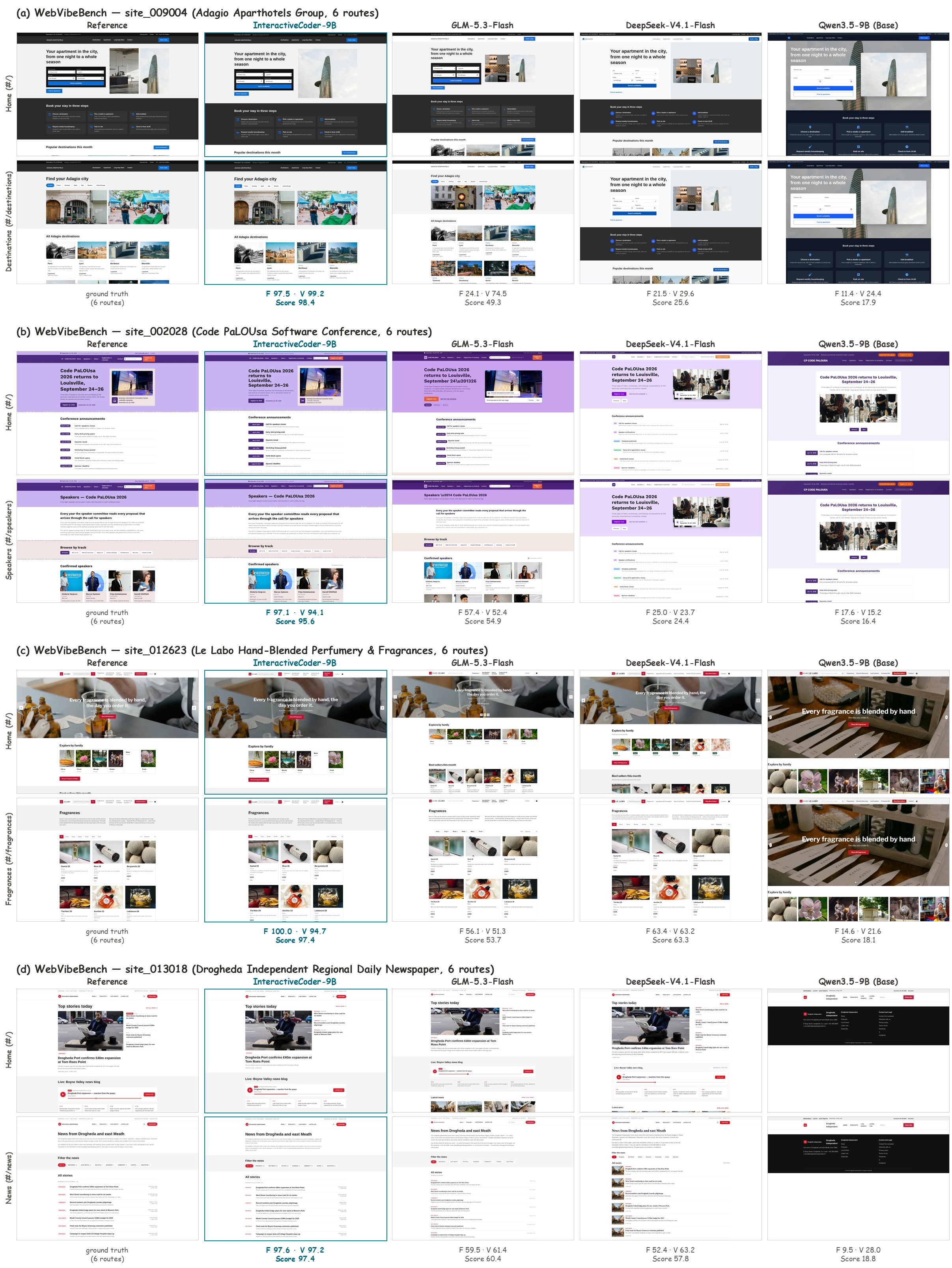}
\caption{\textbf{Additional \arena{} comparisons (2/2).}}
\label{fig:case-wva-app2}
\end{figure*}

\begin{figure*}[t]
\centering
\includegraphics[width=\textwidth,height=0.95\textheight,keepaspectratio]{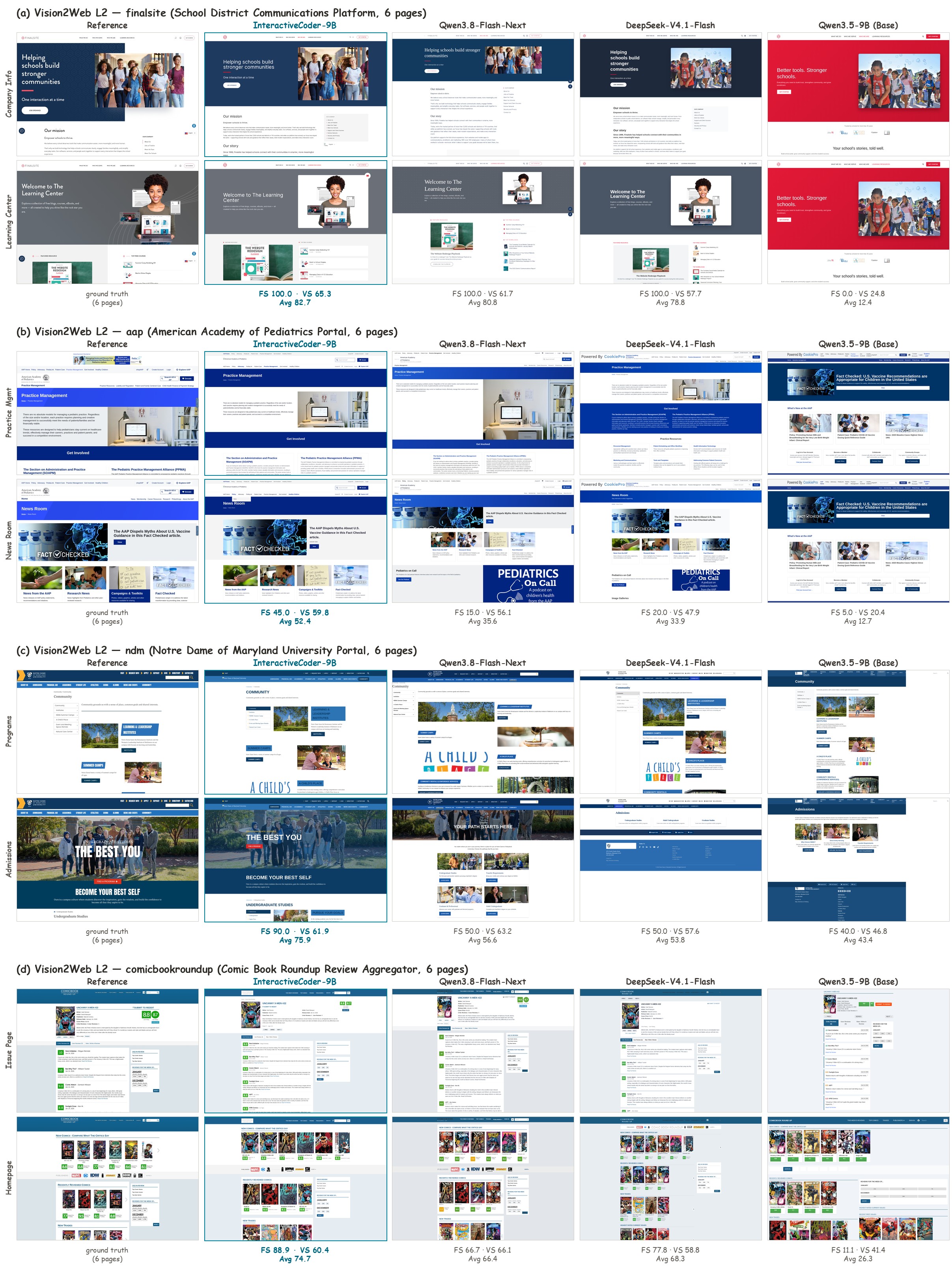}
\caption{\textbf{Vision2Web L2 comparisons (1/2).}}
\label{fig:case-v2w-app1}
\end{figure*}

\begin{figure*}[t]
\centering
\includegraphics[width=\textwidth,height=0.95\textheight,keepaspectratio]{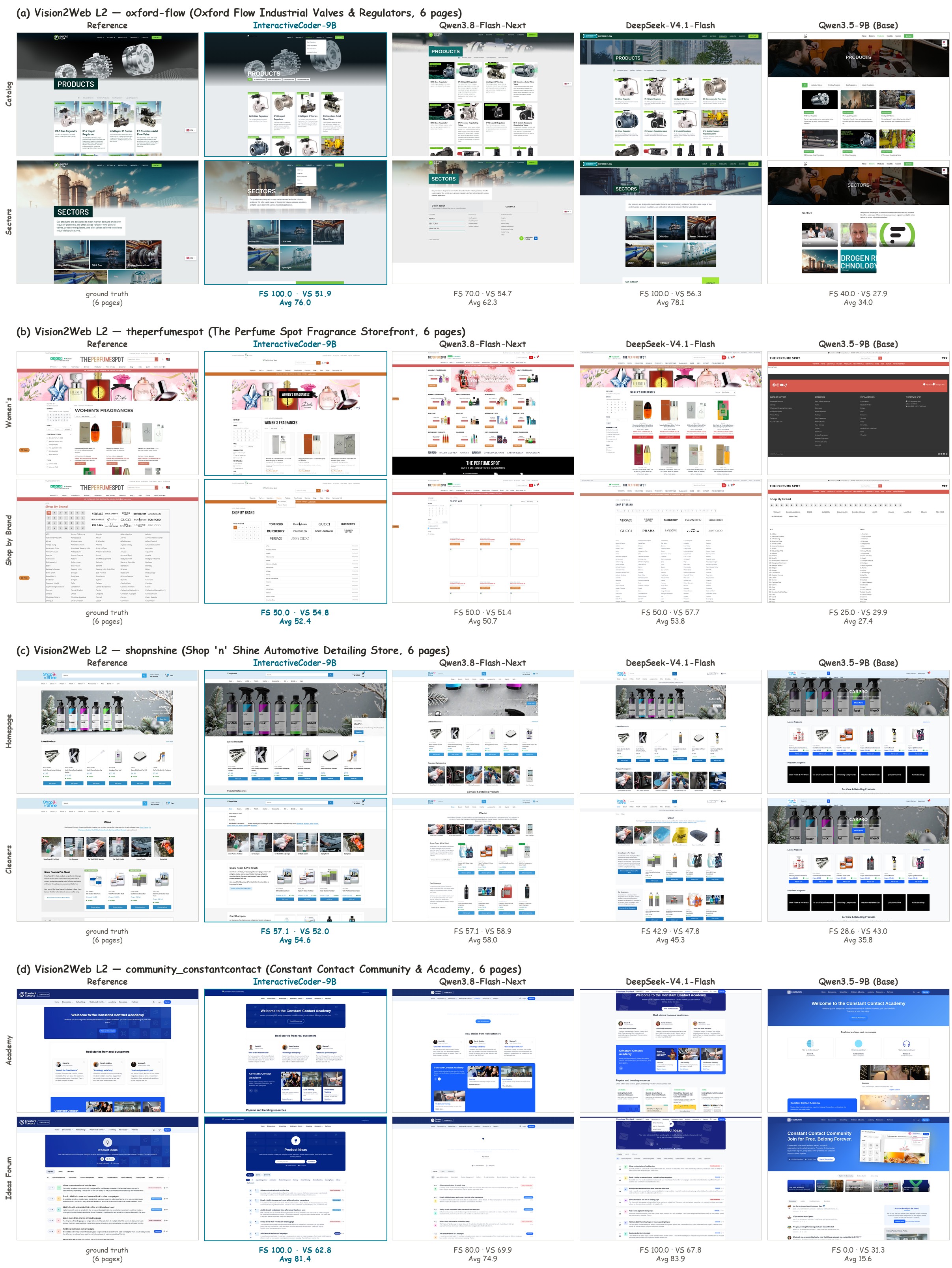}
\caption{\textbf{Vision2Web L2 comparisons (2/2).}}
\label{fig:case-v2w-app2}
\end{figure*}

\clearpage
\onecolumn
\section{Prompts}
\label{app:prompts}

\begingroup
\fontencoding{T1}\selectfont
\newcounter{wvprompt}
\newcommand{\promptref}[1]{Prompt~\ref{prompt:#1}}
\newsavebox{\wvprompthead}
\newsavebox{\wvpromptbody}
\newlength{\wvHh}
\newlength{\wvHb}
\newlength{\wvH}
\newcommand{\promptcard}[3]{%
\refstepcounter{wvprompt}\label{prompt:\thewvprompt}%
\noindent\begin{minipage}{\textwidth}
\vspace{4pt}
\sbox{\wvprompthead}{\parbox{\dimexpr\textwidth-14pt\relax}{\vspace{4pt}\footnotesize\fontencoding{T1}\rmfamily\selectfont\textbf{\textcolor{cvprblue!85!black}{#1}}\hfill{\scriptsize\fontencoding{T1}\rmfamily\selectfont\textcolor{black!70}{#2}}\vspace{4pt}}}%
\sbox{\wvpromptbody}{\parbox{\dimexpr\textwidth-14pt\relax}{\vspace{5pt}\scriptsize\fontencoding{T1}\rmfamily\selectfont\raggedright #3\vspace{5pt}}}%
\setlength{\wvHh}{\dimexpr\ht\wvprompthead+\dp\wvprompthead\relax}%
\setlength{\wvHb}{\dimexpr\ht\wvpromptbody+\dp\wvpromptbody\relax}%
\setlength{\wvH}{\dimexpr\wvHh+\wvHb\relax}%
\begin{tikzpicture}[
  every node/.style={inner sep=0pt, outer sep=0pt, anchor=north west}
]
\begin{scope}
  \path[clip, rounded corners=6pt] (0pt,0pt) -- (\textwidth,0pt) -- (\textwidth,-\wvH) -- (0pt,-\wvH) -- cycle;
  \fill[black!2] (0pt,0pt) rectangle (\textwidth,-\wvH);
  \fill[cvprblue!12] (0pt,0pt) rectangle (\textwidth,-\wvHh);
  \draw[cvprblue!60!black, line width=0.4pt] (0pt,-\wvHh) -- (\textwidth,-\wvHh);
\end{scope}
\node at (7pt,0pt) {\usebox{\wvprompthead}};
\node at (7pt,-\wvHh) {\usebox{\wvpromptbody}};
\draw[cvprblue!60!black, line width=0.6pt, rounded corners=6pt] (0pt,0pt) -- (\textwidth,0pt) -- (\textwidth,-\wvH) -- (0pt,-\wvH) -- cycle;
\end{tikzpicture}
\vspace{4pt}
\end{minipage}\par}

\promptcard{Prompt 1: Unified Student Model Input Template (Training and Evaluation)}{Recipient: Student Policy / Evaluated MLLM}{
You are building one self-contained interactive website as a single index.html file.\\[3pt]
Rules:\\
1. Return only the complete index.html, from <!DOCTYPE html> to </html>, in one \texttt{\textasciigrave\textasciigrave\textasciigrave html} fence. No explanation.\\
2. Load React 18, ReactDOM, Babel and Tailwind from resources/vendor/: react.production.min.js, react-dom.production.min.js, babel.min.js and tailwind-browser.js. Font Awesome is resources/vendor/fontawesome/css/all.min.css. No CDN, npm, import or network request.\\
3. Give every page its own hash route. Every navigation changes location.hash, so the browser Back button returns to the previous page.\\
4. Reproduce the screenshots: layout, section order, colors, typography, text and images. Build every section of every page before refining details.\\
5. Use only the asset files listed below. Write the directory in front of the file name, for example resources/images/\{example\_name\}. Do not invent file names.\\
6. When font files are listed, load them with @font-face and match each one to the screenshots using its specimen tile. When none are listed, use system font stacks that look closest to the screenshots.\\
7. Define each repeated list once as a data array and render it with map. Do not write the same markup out by hand for each item.\\[3pt]
\#\# Task description\\
\{description\}\\[3pt]
\#\# Asset files\\
\{asset\_groups\}\\[3pt]
\#\# Images\\
The following \{image\_count\} images are provided in this order: \{image\_order\}. Screenshots are full pages captured at a \{viewport\_width\}px wide viewport. Asset tiles are labeled with their file names; font files appear as specimen tiles.\\
\{image\_placeholders\}
}

\promptcard{Prompt 2: Style Card Extraction Prompt (Offline Seed Pool)}{Recipient: MLLM Style Analyst (Stage 1)}{
\textbf{[System]} You are a visual design analyst. Describe how a webpage looks so another designer can build a different site in the same visual style.\\[2pt]
Rules:\\
1. Output one JSON object and nothing else. No Markdown fence.\\
2. Describe appearance only. Never copy brand names, headings, body text, product names, people, numbers, addresses or logo shapes from the screenshot. `signature\_details` and `avoid` use generic words ("thin red rule under the header"), never content words.\\
3. Hex colors come from large visible areas: page background, header bar, section bands, primary buttons, body text. Do not guess colors of tiny icons.\\
4. Fonts are described by look, not by name: classify headings and body separately.\\
5. Count what is visible. `section\_count` is the number of distinct vertical blocks from header to footer, counting the header and footer. `capture` is `full\_page` when header and footer are both visible, else `viewport`.\\
6. If the screenshot is mostly a modal, error or blank page, return \{"status": "unusable"\}.\\[2pt]
Schema: \{"status": "ok", "capture": "full\_page | viewport", "page\_kind": "corporate | catalog | storefront | news | magazine | forum | docs | app\_dashboard | portfolio | event | public\_service | education | listing", "palette": \{"page\_bg": "\#hex", "surface": "\#hex", "ink": "\#hex", "muted\_ink": "\#hex", "brand": "\#hex", "brand\_ink": "\#hex", "accent": "\#hex", "band\_colors": ["\#hex"]\}, "header": \{"layout": "...", "background": "...", "nav\_style": "...", "has\_dropdowns": true, "has\_search": false, "cta\_in\_header": true\}, "hero": "full\_bleed\_photo\_overlay | split\_text\_image | carousel | text\_only\_band | no\_hero | product\_grid\_first", "typography": \{"heading\_class": "...", "body\_class": "...", "heading\_case": "...", "heading\_weight": 400, "body\_px": 15, "looks\_like\_system\_font": false\}, "shape": \{"radius\_px": 0, "shadow": "none | subtle | strong", "borders": "none | hairline | strong"\}, "density": \{"section\_count": 9, "content\_width\_px": 1200, "typical\_columns": 3, "text\_density": "sparse | medium | dense", "image\_share": "low | medium | high"\}, "cards": "...", "footer": \{"background": "...", "columns": 4, "has\_newsletter": false, "has\_social\_icons": true\}, "signature\_details": ["at most five short generic traits"], "avoid": ["at most three looks that would break this style"]\}\\[2pt]
\textbf{[User]} Describe the visual style of this screenshot. The page was captured at a \{viewport\_width\}px wide viewport. <image>
}

\promptcard{Prompt 3: Asset Photo Subject Verification Prompt}{Recipient: MLLM Photo Judge (Stage 4)}{
\textbf{[System]} You check stock photos picked by keyword for a website. Each attached sheet shows numbered tiles; the user message gives the subject each tile was picked for.\\
Drop a tile when it is mostly text or graphics rather than a photograph: a slide, poster, flyer, advert or banner with large lettering, a document or form, a screenshot, an infographic, a chart or diagram, a book or product cover dominated by its title, or a collage. Also drop it when it clearly shows something other than its subject.\\
Keep real photographs that plausibly fit the subject, even when not perfect. Small text that belongs to the scene, such as a shop sign or a label, is fine.\\
Reply with JSON only: \{"tiles": [\{"n": 1, "keep": true, "why": ""\}]\} (`why` is a few words, required when `keep` is false).\\[2pt]
\textbf{[User]} Subjects by tile number: \{tiles\_json\}
}

\promptcard{Prompt 4: Site Plan Synthesis Prompt}{Recipient: LLM Site Planner (Stage 2)}{
\textbf{[System]} You are the information architect for a real organisation's website. Plan the site the way that organisation would actually publish it, with its own section names, its own catalogue and its own wording.\\[2pt]
Rules:\\
1. Output one JSON object and nothing else. No Markdown fence.\\
2. Rebuild the website of the real organisation in the seed's `real\_site`. `brand` is its real name. Text carries concrete facts the way a real site does: counts, dates, opening hours, prices, addresses, phone numbers, named people, model numbers. All visible text is in the locale language.\\
3. Page titles and nav labels are the organisation's own words ("Products", "Sectors", "Used Trucks", "Race Info", "Case Studies"). Do not use generic app words such as Browse, Details, Catalog, Saved, Compare, Activity, Overview, Updates unless this organisation would really use them.\\
4. Pages: Exactly `page\_count` top-level pages (first is `home`). `target\_height\_px` of page i is `page\_heights\_px[i]`. Plan at least `minimum\_sections` sections per page (one section per 650px, at least 2 per page). Every non-hero section supplies `minimum\_content\_units\_per\_section` distinct concrete records (cards, rows, features, dates, people, places or FAQ entries) or `minimum\_paragraphs\_per\_text\_section` real paragraphs in `paragraph\_words` range. `image\_subjects`: 1--3 word stock-photo noun tags of a real person, place or object (never charts, UI mockups or icons). Give 2--4 sections a `cta` (\{"label": "", "to": ""\}); at least one on home links to another page.\\
5. Collections: At least `collection\_count` collections, each listed on `listing\_page` with `collection\_items[i].items` distinct items (never numbered like "Item 1" or consecutive codes). Collections with `detail: true` define `detail\_sections` and per-item detail pages reached from cards, related links, search results or breadcrumbs.\\
6. Navigation: `nav` lists top-level pages; when `dropdowns` is true, at least one entry has 2--5 `children` linking to pages or detail pages. `header\_cta` links to a page. `footer` has `footer\_columns` link columns. Every page is reachable from home.\\
7. In-page interactions: at most `in\_page\_count` of `in\_page\_kinds` (`filter | tabs | accordion | carousel | search | form`), each naming its page, exact visible `controls`, `query` (for search), and visible `result`. Do not mention routes, test ids, frameworks, datasets or evaluation in any visible text field.\\[2pt]
Schema: \{"brand": "", "tagline": "", "locale": "en", "about": "", "pages": [\{"id": "home", "title": "", "purpose": "", "target\_height\_px": 4200, "sections": [\{"id": "", "kind": "...", "heading": "", "shows": "", "collection": null, "image\_subjects": [], "icons": [], "logos": [], "cta": null\}]\}], "collections": [\{"id": "", "label": "", "listing\_page": "", "detail": true, "fields": [], "items": [...], "detail\_sections": [...], "related": true\}], "nav": [...], "header\_cta": \{...\}, "footer": \{...\}, "in\_page": [\{"page": "", "kind": "...", "controls": [""], "query": "", "result": ""\}]\}\\[2pt]
\textbf{[User]} Plan this site.\ \ Seed: \{seed\_json\}\ \ Length requirements: \{length\_plan\_json\}\ \ Visual style: \{style\_card\_json\}
}

\promptcard{Prompt 5: Single-File Application Synthesis Prompt}{Recipient: LLM Code Generator (Stage 5)}{
\textbf{[System]} You are a senior front-end engineer, highly sensitive to details and interaction experience and proficient in React and Tailwind CSS, rebuilding a real organisation's website as one interactive single-file app.\\[2pt]
-- \textbf{Runtime:} React 18, ReactDOM 18, Babel (`<script type="text/babel">`), Tailwind v4 (`tailwind-browser.js`), and Font Awesome from `resources/vendor/`. Put element resets inside `@layer base \{ ... \}`. No npm, import, module, fetch, CDN, remote image, `Math.random()`, or clock.\\
-- \textbf{Code shape and size:} Put all content in one `const SITE = \{...\}` object near the top. Render lists with `map` over `SITE` data; one component per section kind and one detail template per collection. Implement at least `length\_plan.minimum\_sections` complete sections per page with distinct real records or substantive paragraphs. Run without any JavaScript error on every route.\\
-- \textbf{Routing and navigation:} Hash router (`\#/` is home) using exact `href` values from the plan on `<a href="\#/...">` so browser Back works. Every non-home page starts `<main>` with one `<h1>` showing the page `title` verbatim (detail page `<h1>` is item `name`). Header nav from `SITE.nav` verbatim; dropdown children open on hover and focus. Logo links to `\#/`; header button uses `header\_cta` label; detail pages include breadcrumbs labeled "Home" and the listing page title.\\
-- \textbf{In-page interactions:} Implement each `in\_page` entry with its exact control labels and visible result: tabs and filters change shown items; accordions open/close; carousels have previous/next controls; search filters items on input/submit; forms validate and show an on-page confirmation.\\
-- \textbf{Appearance, fonts, and images:} Follow the style card hex palette, layout, density, and target heights at 1920px width. Declare one `@font-face` per assigned font face (or system stacks when empty). Reference images strictly via slot tokens (`\_\_PHOTO\_07\_\_`, `\_\_ICON\_02\_\_`, `\_\_LOGO\_01\_\_`) as plain string literals. Return the HTML document only, from `<!DOCTYPE html>` to `</html>`.\\[2pt]
\textbf{[User]} Build this site.\ \ Site plan: \{site\_plan\_json\}\ \ Style card: \{style\_card\_json\}\ \ Image slots: \{slot\_plan\_json\}\ \ Fonts: \{fonts\_json\}
}

\promptcard{Prompt 6: Targeted Search-and-Replace Repair Prompt}{Recipient: LLM Code Generator (Stages 6--8 Repair)}{
\textbf{[System]} (Same system prompt as Prompt 5.)\\[2pt]
\textbf{[User]} Automatic checks ran your page in a browser at 1920px and found these problems:\\
\{errors\}\ \ \{hints\}\\[2pt]
Fix every problem by editing the page. Reply with edit blocks only, in this exact form, as many as needed:\\
\texttt{<\kern0pt<\kern0pt<\kern0pt<\kern0pt<\kern0pt<\kern0pt< SEARCH}\\
\texttt{lines copied exactly from the current page, enough to be unique}\\
\texttt{=======}\\
\texttt{the new lines}\\
\texttt{>\kern0pt>\kern0pt>\kern0pt>\kern0pt>\kern0pt>\kern0pt> REPLACE}\\
Each SEARCH part must match the current page character for character, including indentation. Keep the plan's labels, hrefs, slot tokens and design everywhere else.
}

\promptcard{Prompt 7: Per-Page and Cross-Page Visual Quality Audit Prompt}{Recipient: MLLM Visual Judge (Stage 9)}{
\textbf{[System]} You are a strict quality-control judge for synthetic interactive websites. For each attached screenshot, mark that page `drop` if any defect is present:\\
-- `broken\_image`: broken-image icon, visible alt text, empty image frame, or placeholder where a real image should be.\\
-- `broken\_layout`: severe overlap, unreadable text collision, major content outside the viewport, horizontal overflow, or collapsed composition.\\
-- `blank`: mostly empty/solid page with too little rendered content.\\
-- `blocked`: access/login/subscription/cookie/verification overlay obscures the main content.\\
-- `non\_webpage`: essentially one poster/image/video cover without webpage structure.\\
-- `image\_text\_mismatch`: prominent imagery is clearly unrelated to the named page or its visible text.\\
Also judge whether the pages clearly belong to the same product. Mark `cross\_page` as `drop` only for `cross\_page\_inconsistency`: shared navigation, typography, colors, spacing, or component language vary so severely that the pages look like unrelated websites.\\
Return exactly one JSON object: \{"pages": [\{"verdict": "keep | drop", "file": "00\_home.png", "reasons": [], "offending\_slots": [], "explanation": "..."\}], "cross\_page": \{"verdict": "keep | drop", "reasons": [], "explanation": "..."\}\}.\\[2pt]
\textbf{[User]} Scene and page context: \{context\_json\}\ \ Attached screenshots, in order: \{screenshots\_json\}\ \ Inspect every attached image and return the JSON verdict.
}

\promptcard{Prompt 8: Reverse Task Description Writer Prompt}{Recipient: LLM Description Writer (Stage 10)}{
\textbf{[System]} You write the task description for a website build request, in the voice of the site owner describing how visitors move through the site.\\[2pt]
Rules:\\
1. Plain prose only. No title, Markdown, bullets, numbering, JSON or code.\\
2. Open with `I want to build` naming the brand and the kind of website, walk from page to page in first-person voice, and close with the shared header, logo and footer.\\
3. Cover this site's homepage and the verified journeys: what I click, where it takes me, what that page shows, and what I can do from there. Mention dropdown children, call-to-action buttons, footer links, breadcrumbs and related-item cards where a journey uses them.\\
4. Describe each in-page interaction by what a visitor sees change. Never describe an assertion, a count check or a test.\\
5. Use only supplied facts. Quote visible labels in double quotes (about 10--35 quoted labels, at most \{max\_quotes\}).\\
6. Never mention routes, hashes, slugs, ids, selectors, frameworks, code, screenshots, prototypes, datasets or evaluation. Stay within \{min\_words\}--\{max\_words\} words.\\[2pt]
\textbf{[User]} Site plan: \{site\_plan\_json\}\ \ Grounded facts from the rendered pages: \{facts\_json\}\ \ Verified journeys and in-page interactions: \{verified\_json\}\ \ Write the task description.
}

\promptcard{Prompt 9: Block-Wise Visual Similarity Judge Prompt (RL Reward and Evaluation)}{Recipient: MLLM Visual Judge ($v_{\text{judge}}$)}{
You are a senior QA automation engineer. Your task is to compare a prototype image with an actual page screenshot and score the appearance of the page. Carefully observe all differences between the prototype and the actual page.\\[2pt]
\#\# Input:\ \ 1. Prototype Image: <|prototype\_page|>\ \ \ 2. Actual Page Image: <|actual\_page|>\\[2pt]
\#\# Component Segmentation Rules:\\
Divide the page into logical functional blocks (Navigation bar / header, Hero section / main banner, Product listing / catalog area, Forms / sidebar, Footer, and any other visually distinct functional sections). Treat each block as a single component for scoring.\\[2pt]
\#\# Scoring Rules (per component):\\
-- \textbf{1.0 (Perfect Match):} Position, layout, spacing, alignment, sizing, text, fonts, colors, icons, and images are identical.\\
-- \textbf{0.75 (Minor Imperfections):} Position mostly accurate (<2px misalignment); layout and text largely consistent; slight scaling or color variations noticeable only on close inspection.\\
-- \textbf{0.5 (Partial Match):} Position roughly correct but noticeable misalignment or spacing issues; several text discrepancies; media partially incorrect or missing.\\
-- \textbf{0.25 (Poor Match):} Recognizable but significantly misaligned; layout mostly inconsistent; text differs significantly; media missing or wrong.\\
-- \textbf{0.0 (No Match):} Component missing, completely misplaced, or unrecognizable.\\[2pt]
\#\# Output Requirements: Output only a JSON array: [\{"name": "<component name>", "score": <0 | 0.25 | 0.5 | 0.75 | 1>, "reason": "<brief explanation>"\}]
}
\endgroup

\fi

\ifwvappendix
  \ifwvmain\else
    \clearpage
    \twocolumn
    \wvprintbibliography
  \fi
\fi

\end{document}